\documentclass[journal]{IEEEtran}
\usepackage{amsmath,amsfonts}
\usepackage{algorithmic}
\usepackage{algorithm}
\usepackage{array}
\usepackage{textcomp}
\usepackage{stfloats}
\usepackage{url}
\usepackage{color}
\usepackage{verbatim}
\usepackage{graphicx}
\usepackage{cite}
\usepackage{bm}
\usepackage{hyperref}
\usepackage[table]{xcolor}
\definecolor{mygray}{gray}{.92}
\usepackage{subfigure}
\usepackage{tabularx}
\begin{document}

\title{Hierarchical Semantic Compression for Consistent Image Semantic Restoration}

\author{Shengxi Li,~\IEEEmembership{Member,~IEEE,} Zifu Zhang,~\IEEEmembership{Student Member,~IEEE,} Mai Xu,~\IEEEmembership{Senior Member,~IEEE,}\\ Lai Jiang, \IEEEmembership{Member,~IEEE,} Yufan Liu, \IEEEmembership{Member,~IEEE,} Ce Zhu,~\IEEEmembership{Fellow,~IEEE}\\\vspace{-2em}
        % <-this % stops a space
\thanks{Shengxi Li, Zifu Zhang, Lai Jiang and Mai Xu are with the School of Electronic and Information Engineering, Beihang University, Beijing 100191, China (Email: \{LiShengxi; ZifuZhang; Jianglai.china; MaiXu\}@buaa.edu.cn). Yufan Liu is with State
 Key Laboratory of Multimodal Artificial Intelligence Systems, Institute of Automation, Chinese Academy of Sciences, Beijing, 100190, China (Email: yufan.liu@ia.ac.cn). Ce Zhu is with the School of Electronic and Information Engineering, University of Electronic Science and Technology of China, Chengdu 611731, China (Email: eczhu@uestc.edu.cn).}% <-this % stops a space
}

\markboth{Journal of \LaTeX\ Class Files,~Vol.~14, No.~8, August~2021}%
{Shell \MakeLowercase{\textit{et al.}}: A Sample Article Using IEEEtran.cls for IEEE Journals}


\maketitle

\begin{abstract}
The emerging semantic compression has been receiving increasing research efforts most recently, capable of achieving high fidelity restoration during compression, even at extremely low bitrates. However, existing semantic compression methods typically combine standard pipelines with either pre-defined or high-dimensional semantics, thus suffering from deficiency in compression. To address this issue, we propose a novel hierarchical semantic compression (HSC) framework that purely operates within intrinsic semantic spaces from generative models, which is able to achieve efficient compression for consistent semantic restoration. More specifically, we first analyse the entropy models for the semantic compression, which motivates us to employ a hierarchical architecture based on a newly developed general inversion encoder. Then, we propose the feature compression network (FCN) and semantic compression network (SCN), such that the middle-level semantic feature and core semantics are hierarchically compressed to restore both accuracy and consistency of image semantics, via an entropy model progressively shared by channel-wise context. Experimental results demonstrate that the proposed HSC framework achieves the state-of-the-art performance on subjective quality and consistency for human vision, together with  superior performances on machine vision tasks given compressed bitstreams. This essentially coincides with human visual system in understanding images, thus providing a new framework for future image/video compression paradigms. Our code shall be released upon acceptance.
\end{abstract}


\section{Introduction}
\IEEEPARstart{R}{ecent} {advances on multimedia services have been popularising the visual-based content in various applications and leading to an explosive increase on image/video data volume, thus calling for efficient image compression methods to relieve the huge demand on bandwidth and storage resources. Since the first image compression standard {JPEG\cite{hudson2018jpeg}} proposed in {1980s}, successive standards have been released including {H.264/AVC \cite{wiegand2003overview}, H.265/HEVC \cite{sullivan2012overview}, H.266/VVC \cite{bross2021overview}}, each of which sequentially improves the coding efficiency  at the cost of significantly increased computational complexity \cite{hussain2018image}. The past decade has also witnessed a parallel path to improve the coding efficiency by using deep learning techniques, in which encoder, decoder, entropy models, etc., are built upon the deep neural networks {\cite{jamil2023learning, balle2017end, balle2018variational, cheng2020learned}}. However, the majority of image compression methods aim to improve the pixel-wise accuracy, and gradually encounter the bottleneck that further improving the coding efficiency typically requires sophisticated designs and excessive computational complexity \cite{hu2021learning}. } 

\begin{figure}[t]
   \begin{center}
   \includegraphics[width=0.96\linewidth]{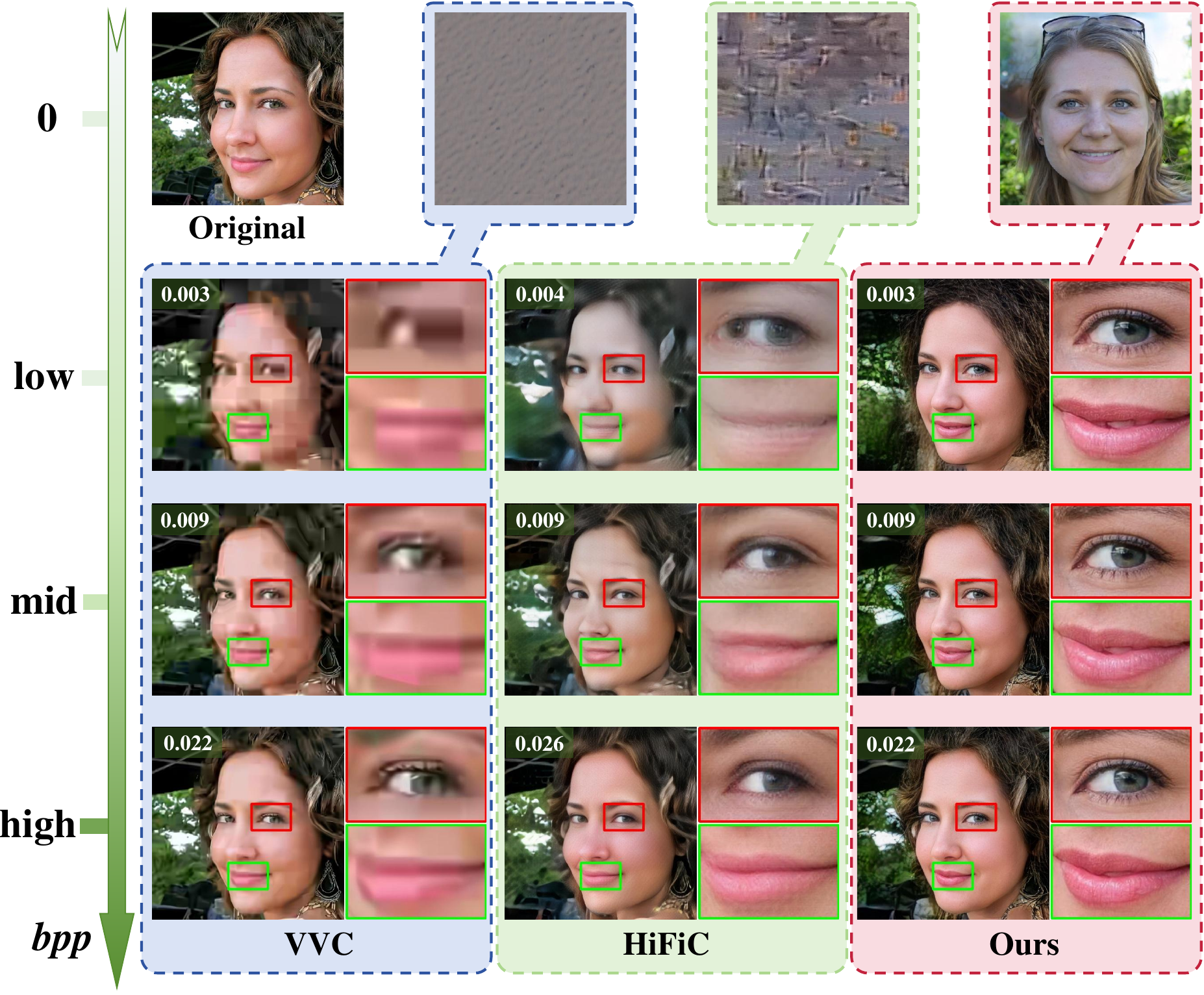}
   \end{center}
   \vspace{-1.5em}
      \caption{{Illustration of restoring image semantic consistency instead of pixel-wise accuracy, by representative methods including VVC for the latest standard codec, HiFiC for the subjective-oriented compression and our HSC for the semantic compression. Note that our HSC method achieves continuous bitrates starting from 0 bit per pixel (bpp), by compressing intrinsic semantics.} }
   \label{fig:sc}
   \vspace{-1.5em}
\end{figure}

{Since human visual system is the final receiver for most image compression scenarios, the way to improve the coding efficiency is encouraged to enhance the subjective quality perceived by human {\cite{blau2019rethinking}}, rather than the objective quality that is dominantly reflected by pixel-wise accuracy. Correspondingly, various subjective-driven methods, such as weighted resource allocation {\cite{li2017closed, jiang2022does}} and perceptual losses {\cite{blau2018perception, mentzer2020high}}, have been applied in addition to solely optimizing pixel-wise losses; this way, the coding procedure is under the guidance from human region of interest (ROI) both implicitly and explicitly. Most recently, semantic compression has been receiving increasing research efforts to guarantee superior subjective quality, by means of condensing and transmitting pre-defined semantics of images, such as textural \cite{wang2021towards, arikan2024semantic} and {structural} \cite{prakash2017semantic, yan2021sssic} cues; this approximates the behaviour when human communicate, thus capable of achieving extremely low bit-rate compression towards human visual perception and understanding. However, existing methods extract either pre-defined or high-dimensional semantics in an \textit{ad hoc} way, in which the effectiveness is restricted in certain scenarios and formats. }

{Instead of pre-defined semantics, deep generative models have been verified to possess rich intrinsic semantics in the noisy latent generating space \cite{shen2020interpreting, kwon2022diffusion}.  Compared to diffusion models, the generative adversarial network (GAN) essentially provides a well-defined basic latent space for compressing semantics within images, which is able to generate highly realistic images from merely small latent dimensions. Based on GANs, one of the main obstacles regarding semantic compression lies on the unidirectional generation from latent space to the image side, whereas embedding real-world images into the latent space is typically intractable for the majority of GANs. Most recently, an encoder was employed to embed images into the style code features of StyleGAN, which are then compressed for face images. The style code features  possess rich semantics with significantly reduced compressing dimensions, such that the compression enables human-machine collaborative vision tasks \cite{mao2023scalable}. However, solely compressing the semantic features cannot ensure the accurate reconstruction on details, in which the existing method is still limited on face image scenarios. Indeed, given pre-trained GANs, accurately reconstructing the details typically exhibits the trade-off against precisely restoring the semantics,  which also corresponds to the well-known rate-distortion-editing trade-off in the compression \cite{wang2021HFGI}. }

\begin{figure*}[htbp]
   \begin{center}
       \includegraphics[width=0.96\linewidth]{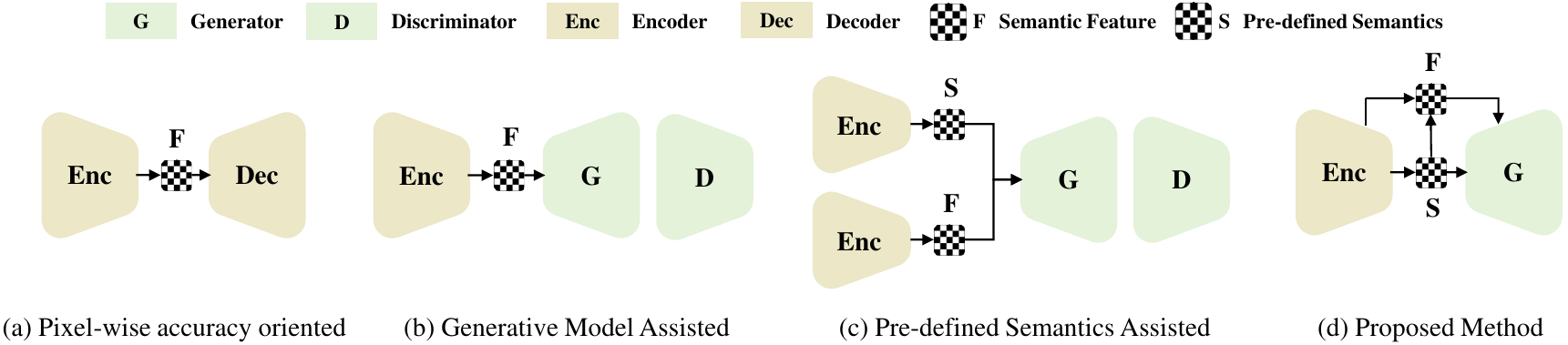}
   \end{center}
   \vspace{-1.5em}
      \caption{{Main categories of existing image compression paradigms. (a) denotes image compression that aims to optimise pixel-wise accuracy, including VVC \cite{bross2021overview} and many learned methods \cite{balle2017end, balle2018variational, he2022elic}. (b) represents image compression assisted by generative models to improve perceptual quality, including HiFiC \cite{mentzer2020high} and MS-ILLM \cite{muckley2023improving}. (c) depicts existing semantic compression methods that almost are built upon pre-defined semantics, such as texts  \cite{lei2023text+sketch} and structures \cite{chang2022conceptual}. (d) represents the proposed HSC method, which is purely based on the intrinsic semantics from unconditional advanced generative models.}}
   \label{fig:introduction}
   \vspace{-1em}
\end{figure*}

{In this paper, we propose a novel hierarchical structure to compress intrinsic semantic features at different stages with shared information, such that the details can be accurately recovered whilst the semantics is consistently retained. More specifically, we first analyse the general framework of image compression via the entropy relationships between images and compressing latent features. The analysis reveals the benefits of employing an advanced deep generative model, and thus motivates us to build our hierarchical semantic compression framework (HSC) based on the StyleGAN architecture, which is able to achieve state-of-the-art generation performances \cite{yao2022style}.  Then, we develop a general inversion encoder to address the unidirectional generation problem of StyleGAN, such that the middle-level semantic feature and core semantics can be recovered given pre-trained StyleGAN. Correspondingly, the middle-level semantic feature is compressed by a newly proposed feature compression network (FCN), with the aim to retain the accuracy of restored details during compression. On the other hand, we propose a semantic compression network (SCN) to compress the core semantics, such that the semantic consistency is guaranteed. The FCN and SCN operate in a hierarchical style, in which a progressive shared entropy model is developed to further reduce the semantic redundancy across features. This way, our HSC method is able to achieve both semantic consistency and restoration accuracy for compression, in which we further develop a semantic editing pipeline that allows for straightforward editing based on compressing bitstreams, a natural outcome of our semantic compression paradigm. Consequently, experimental results verify the superior performances of our HSC method, on subjective quality and semantic consistency for human visions, together with the state-of-the-art accuracy on machine vision tasks. }

\section{Related Works}

{Benefiting from the universal approximation capability, deep neural networks (DNNs) have been acting as the workhorse in existing learnt image compression methods. Thus, we first review the methods that dominantly focus on the pixel-wise accuracy, followed by the methods for optimising the perceptual quality assisted by generative models. Regarding the emerging trends on compressing pre-defined semantics, we finally review the semantic compression methods. We visualise their distinct difference and highlight our novelty  in Fig. \ref{fig:introduction}.}

\subsection{Pixel-wise Accuracy Oriented Compression}

{Indeed, existing standard compression codecs evolved by consistently improving the mean squared error (MSE), including JPEG \cite{marcellin2000overview}, H.264/AVC \cite{wiegand2003overview}, H.265/HEVC \cite{sullivan2012overview} and H.266/VVC \cite{bross2021overview}. During the past decade, DNN-based image compression has achieved remarkable progress by jointly optimizing the rate-distortion performance, in which the pixel-wise accuracy was dominantly considered by the distortion metrics. To the best of our knowledge, Toderici \textit{et al.} proposed the first end-to-end image compression algorithm in 2015, which developed a recurrent neural network to iteratively encode image residuals \cite{toderici2015variable}; this method has surpassed JPEG when compressing small-resolution images. Benefiting from the success of convolutional neural networks (CNNs), Ball{\'e} \textit{et al.} \cite{balle2017end} proposed an end-to-end CNN-based image compression framework, significantly stabilizing and improving image compression efficiency. Follow-up improvements include the hyper-prior structure to de-correlate latent space \cite{balle2018variational}, auto-regressive model to predict context information \cite{minnen2018joint}, attention-inspired approach \cite{cheng2020learned} and invertible neural networks \cite{xie2021enhanced} to improve the feature transformation. The above methods are optimized mostly against the objective quality, e.g., minimizing the MSE or multi-scale structure similarity index measure (MS-SSIM) \cite{wang2004image,wang2003multiscale} of reconstructed images, which now have outperformed the state-of-the-art standard VVC codec regarding pixel-wise accuracy \cite{he2022elic}. However, the pixel-wise accuracy, although improved by advanced metrics such as human region of interest (ROI) weighted MSE and MS-SSIM \cite{wang2003multiscale}, are still yet to match human subjective perception \cite{kim2017deep}, in which a visually inperceptible misalignment on images can lead to remarkable degradation on those pixel-wise metrics. }

\subsection{Generative Model Assisted Compression}

{Since human perception is the ultimate end of restored images, improving the subjective quality of reconstructed images is thus the long-standing necessity in the field of image compression. Benefiting from the most recent advances of deep generative models in outputting realistic images, we have been witnessing compression methods to incorporate GANs to improve subjective quality by learning similarity across distributions, apart from minimizing the pixel-wise metrics. Rippel \textit{et al.} \cite{rippel2017real} employed the adversarial learnt discriminator to remove the block effect of compressed images. Then, Agustsson \textit{et al.} \cite{agustsson2019generative} investigated both unconditional and conditional GANs for full-resolution image compression, and demonstrated that using GAN loss with rate-distortion joint optimization can result in significant bitrate savings. Mentzer \textit{et al.} \cite{mentzer2020high} further built upon conditional GAN, with sophisticated designs including modify the normalization function to eliminate darkening artefacts and the hyperprior model to enhance compression efficiency. Subsequent researches also enhanced the encoder-decoder architecture \cite{he2022po}, with local adversarial discriminators \cite{muckley2023improving}, and the conditional generator \cite{agustsson2023multi} to further improve compression efficiency. With the rapid development of diffusion models \cite{ho2020denoising,song2020denoising}, diffusion-based autoencoder models have been also investigated to reconstruct images with high fidelity \cite{preechakul2022diffusion}. Based on this, Yang and Mandt \cite{yang2024lossy} employed the encoder to map images into a quantised contextual latent variable, which then aids the diffusion model as the decoder to yield competitive compression performance with GAN-based models. However, the above methods, in which the subjective quality is improved with the aid of deep generative models, still encounter the trade-off between the objective and subjective quality when optimizing the compression performance. }

\subsection{Semantic Image Compression}

{Human communicate by linguistic semantics, which motivates the most recent research on semantic compression to focus on restoring the semantic information instead of the pixel-wise details of images. This allows for compressing images at extremely low bit-rates. Existing semantic compression methods typically rely on pre-defined semantics, including learnt structural representations \cite{zhang2024machine}, semantic maps \cite{huang2021deep, korber2024egic, akbari2019dsslic}, texture maps \cite{chang2023semantic} and text \cite{lee2024neural}, to improve the subjective consistency of the reconstructed images. We also noticed that multiple forms of pre-defined semantics are employed to improve compression efficiency. Chang et al. \cite{chang2022conceptual} proposed to compress images by a dual-layered model, consisting of structural layer and texture layer to achieve improve bitrate-quality performances.  Li et al. \cite{li2021cross} also proposed a  cross-modal compression framework based on the text, sketch and semantic maps, to transform the redundant visual data into a compact and human-comprehensible domain. However, the above methods still derive semantic information  from images via pre-defined styles; this is still \textit{ad hoc} to summarise compact and precise image semantics for compression. }

{Most recently, with the rapid development of large-scale models, several methods \cite{lei2023text+sketch, careil2023towards, li2024misc} were built upon pre-trained text-conditioned diffusion models, to benefit the compression from the semantics embedded in the models. Those advanced methods, however, still rely on compressing the extra text or sketch information as auxiliary cues, which suffers from the deficiency on bitrates. The latent generating space of diffusion models also retains the high dimensions, which exhibits unstructured space that requires careful strategies to alter the semantics \cite{wu2023latent, guo2024smooth}. Rather than the diffusion models that retains feature sizes from the beginning to the end, the latent generating space of GANs typically resides on extremely low dimensions and essentially possesses rich intrinsic semantics, which is highly suitable for the compression task. The most recent work \cite{mao2023scalable} established an encoder to reconstruct the latent style codes for StyleGAN, which achieved superior performances on compressing face images. Its reconstruction accuracy, as well as the application scenarios, are relatively limited. To the best of our knowledge, this paper sets out the first attempt to achieve the state-of-the-art semantic compression, based on GANs for various scenarios.} 

\vspace{-1em}
\section{Methodology}

%{We introduce the proposed HSC method for efficient image compression. Before introducing the details, we first analyse the general framework for semantic compression in Section \ref{sec:method-A}, in particular regarding the entropy relationships. Motivated by this analysis, we introduce our hierarchical architecture in Section \ref{sec:method-B}, especially for the shared entropy model across channels.  In Section \ref{sec:method-C}, we  introduce the overall network architecture of our HSC method, followed by the optimisation strategy in Section \ref{sec:method-D}. We finally introduce the usage of semantics intrinsically retained by our HSC method in Section \ref{sec:method-E}, allowing for various forms of semantic editing.}

\begin{figure*}[htbp]
   \begin{center}
   \includegraphics[width=0.96\linewidth]{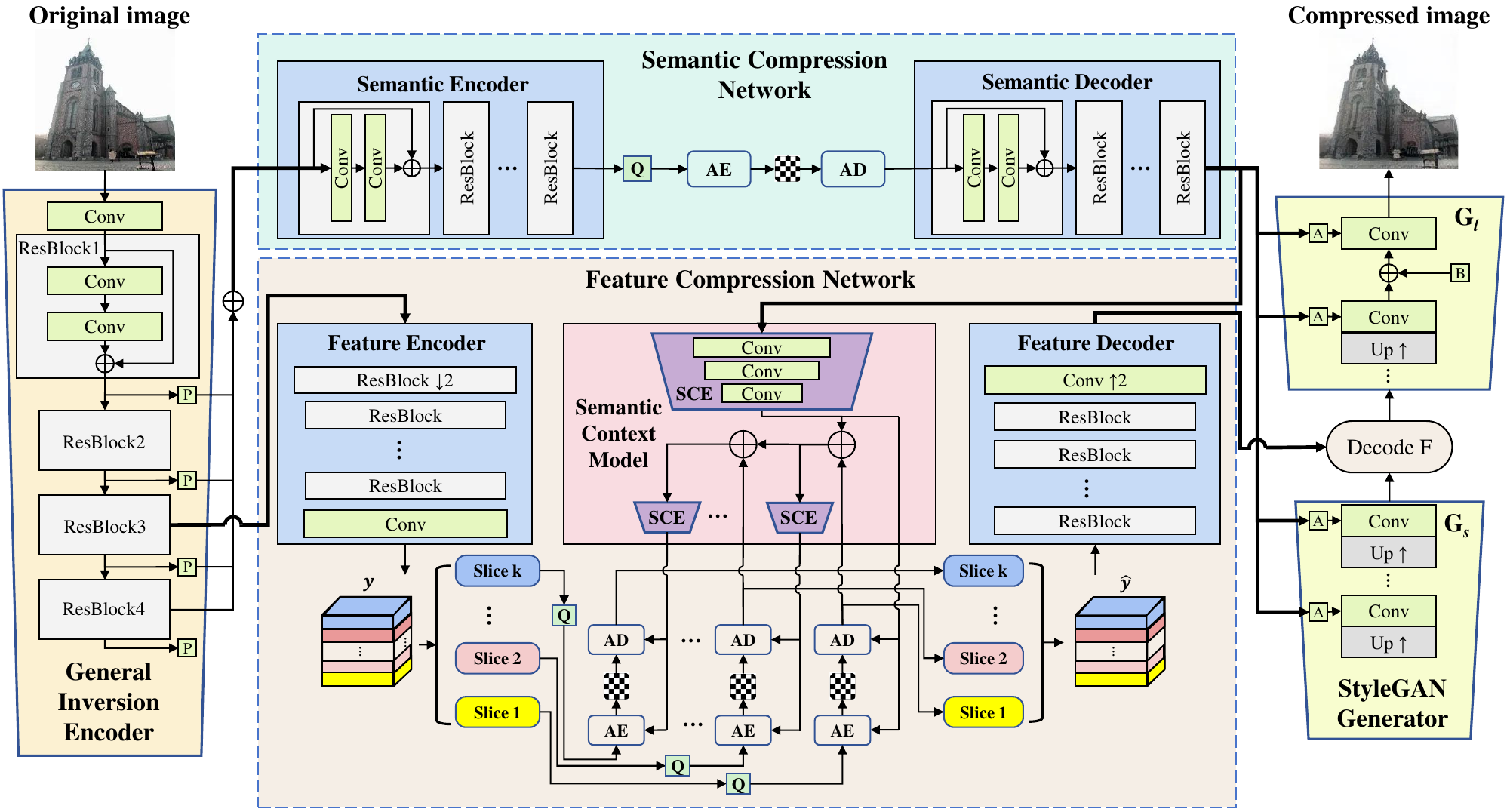}
   \end{center}
   \vspace{-1.5em}
      \caption{{The overall architecture of the proposed HSC
      method, which includes the general inversion encoder (GIE), feature compression network (FCN), semantic compression network (SCN), and pre-trained StyleGAN generator. We separate the StyleGAN generator by the proposed intermediate semantic feature, apart from solely compressing the core semantics from the style codes. This way, the proposed HSC compresses the semantics from the GIE network in a hierarchical way, given the mutual semantics between SCN and FCN during compression, guaranteeing both semantic consistency and high fidelity.}}
   \label{fig:pipeline}
   \vspace{-0.5em}
\end{figure*}

\subsection{Analysis on Semantic Compression}\label{sec:method-A}
{Without lose of generality, existing learnt compression framework typically consists of sequentially composed modules, in which the transformation of each module is represented by $T_i(\cdot)$. Correspondingly, we represent the input and output features by $\mathbf{f}_{i-1}\in \bm{\mathcal{F}}_{i-1}$  and $\mathbf{f}_{i}\in \bm{\mathcal{F}}_{i}$, whereby $i=1,2,\ldots,n$, $\bm{\mathcal{F}}_0=\bm{\mathcal{X}}$ and $\bm{\mathcal{F}}_n$ is the feature to be compressed.  More importantly, to satisfy the aim of compression,  $T_i(\cdot)$ essentially acts as surjective (or bi-jective) from high dimensions to low dimensions; this leads to the conditional entropy $H(\bm{\mathcal{F}}_{i}|\bm{\mathcal{F}}_{i-1})=0$ given two consecutive features $\bm{\mathcal{F}}_{i-1}$ and $\bm{\mathcal{F}}_{i}$. This way, we obtain the following relationship regarding the entropy:
\begin{equation}
\begin{aligned}
     H(\bm{\mathcal{F}}_{i-1})&=H(\bm{\mathcal{F}}_{i-1})+H(\bm{\mathcal{F}}_{i}|\bm{\mathcal{F}}_{i-1})\\
     &= H(\bm{\mathcal{F}}_{i},\bm{\mathcal{F}}_{i-1})=H(\bm{\mathcal{F}}_{i})+H(\bm{\mathcal{F}}_{i-1}|\bm{\mathcal{F}}_{i}).
\end{aligned}
\end{equation}
Then, we are able to represent the entropy of features in a recursive style as follows,
\begin{equation}\label{eq_recur}
\begin{aligned}
    H(\bm{\mathcal{X}})&=H(\bm{\mathcal{F}}_0)=H(\bm{\mathcal{F}}_1) + H(\bm{\mathcal{F}}_0|\bm{\mathcal{F}}_1)\\
    &=H(\bm{\mathcal{F}}_2) + H(\bm{\mathcal{F}}_1|\bm{\mathcal{F}}_2)+H(\bm{\mathcal{F}}_0|\bm{\mathcal{F}}_1)\\
    &=H(\bm{\mathcal{F}}_n) + \sum_{i=1}^{n}H(\bm{\mathcal{F}}_{i-1}|\bm{\mathcal{F}}_i).
\end{aligned}
\end{equation}
We may need to point out that similar relationship is obtained in previous studies \cite{yan2021sssic}, whilst as shall be explained shortly, we reformulate this by different goals.  In \eqref{eq_recur}, $\bm{\mathcal{F}}_n$ has to be compressed with the bit-rate $H(\bm{\mathcal{F}}_n)$. Due to $H(\bm{\mathcal{F}}_{i-1}|\bm{\mathcal{F}}_i) \geq 0$, we may achieve reduced $H(\bm{\mathcal{F}}_n)$ to save bit-rates, when the number of modules $n$ increases. However, the accuracy of reconstructed images may lose due to limited information retained by $\bm{\mathcal{F}}_n$, which highlights the importance of smartly choosing $\bm{\mathcal{F}}_n$ for lossy image compression. Existing learnt compression methods choose either specified tensor \cite{chang2022conceptual} or pre-defined semantics \cite{lei2023text+sketch} as $\bm{\mathcal{F}}_n$, in which the improvement on compression efficiency is still limited.}

{By inspecting \eqref{eq_recur}, we can find that given fixed $H(\bm{\mathcal{X}})$ and $n$,  further reducing $H(\bm{\mathcal{F}}_n)$ is able to be achieved by increasing $\sum_{i=1}^{n}H(\bm{\mathcal{F}}_{i-1}|\bm{\mathcal{F}}_i)$. By learning the complete image distribution, GAN is able to possess maximised $H(\bm{\mathcal{X}}|\bm{\mathcal{S}})=H(\bm{\mathcal{F}}_0|\bm{\mathcal{S}})$ from extremely small dimensions on $\bm{\mathcal{S}}$; this motivates our semantic compression pipeline based on existing state-of-the-art GANs. More specifically,  we seek to develop an encoder that is able to extract the pure semantics $\bm{\mathcal{S}}$ from GANs, and this encoder is thus also the surjective (or bi-jective) mapping from high dimensions to low dimensions. We thus have 
\begin{equation}\label{eq_semantic}
H(\bm{\mathcal{X}})=H(\bm{\mathcal{S}})+H(\bm{\mathcal{X}}|\bm{\mathcal{S}}).
\end{equation}
Therefore, by combining \eqref{eq_recur}, we arrive at 
\begin{equation}\label{eq_mix}
\begin{aligned}
H(\bm{\mathcal{X}})=\alpha\big(&H(\bm{\mathcal{S}})+H(\bm{\mathcal{X}}|\bm{\mathcal{S}})\big)\\
&+(1-\alpha)\big(H(\bm{\mathcal{F}}_n) + \sum_{i=1}^{n}H(\bm{\mathcal{F}}_{i-1}|\bm{\mathcal{F}}_i)\big),
\end{aligned}
\end{equation}
where $\alpha\in[0,1]$ denotes the balance. In \eqref{eq_mix}, the feature $\bm{\mathcal{F}}_n$ and semantics $\bm{\mathcal{S}}$ need to be compressed together, resulting into the minimised bit-rate $\alpha H(\bm{\mathcal{S}})+(1-\alpha)H(\bm{\mathcal{F}}_n)$. Furthermore,  benefiting from the diversifying and complete distribution learnt by GANs, we have $H(\bm{\mathcal{X}}|\bm{\mathcal{S}})>\sum_{i=1}^{n}H(\bm{\mathcal{F}}_{i-1}|\bm{\mathcal{F}}_i)$ and $H(\bm{\mathcal{S}})<H(\bm{\mathcal{F}}_n)$, in which $\alpha H(\bm{\mathcal{S}})+(1-\alpha)H(\bm{\mathcal{F}}_n)<H(\bm{\mathcal{F}}_n)$ for $\alpha>0$. Therefore, involving the rich semantics from GANs in \eqref{eq_mix} is able to further save bit-rates against the existing image compression pipeline in  \eqref{eq_recur}.}

 {Based on above analysis on compressing images with semantics, we propose to compress both the middle-level feature $\bm{\mathcal{F}}_n$  and core semantics $\bm{\mathcal{S}}$ to save bitrates whilst improving the reconstruction accuracy and fidelity.  Benefitting from the intrinsic characteristics of the semantics from generating spaces of GANs, we are able to compress $\bm{\mathcal{F}}_n$ and  $\bm{\mathcal{S}}$ hierarchically with mutually shared cues, different from existing methods that extract pre-defined semantics and compress bit-streams independently. This further removes the redundancy and saves the bitrates of compressed images. Moreover, given the state-of-the-art performances, our HSC method is built upon the StyleGAN architecture \cite{karras2020analyzing}, and the overall architecture is shown in Fig. \ref{fig:pipeline}, which includes general inversion encoder (GIE),  semantic compression network (SCN), feature compression network (FCN), and a pre-trained StyleGAN generator. }
\vspace{-0.5em}
\subsection{Compression on Middle-level Semantic Feature}
\label{sec:method-B}

{Since the latent style code $\mathbf{w}\in\bm{\mathcal{W}}$ of StyleGAN possesses rich and complete semantics \cite{xia2022gan} as the semantics, we define our semantics $\bm{\mathcal{S}}$ as $\bm{\mathcal{W}}$, which includes $m$ latent style codes $\mathbf{w}$ that are able to implicitly control the objects, colour, orientation, textures, to name but a few \cite{xia2022gan}.  Furthermore, we employ symmetric architecture to reconstruct the intermediate feature $\bm{\mathcal{F}}_n$ given the generator of StyleGAN, such that the detailed semantics can be well aligned and restored. Therefore, the proposed GIE network, consisting of residual block modules, aims to restore the multi-stage semantics from the generator, which can be formulated as
\begin{equation}\label{eq_sf}
\begin{aligned}
    \mathbf{f} &= \mathrm{GIE}_n[\mathbf{x}]\\
\mathbf{s}_j&=\mathrm{Lin}_j\big[\bigoplus_{i=1}^{N}p(\mathrm{GIE}_i[\mathbf{x}])\big],
\end{aligned}
\end{equation}
where $\mathbf{x}$ denotes the images to be compressed, $\mathbf{f}\in\bm{\mathcal{F}}_n$ denotes the middle-level semantic feature obtained from the $n$-th GIE module (denoted by $\mathrm{GIE}_n[\cdot]$), and $\mathbf{s}_j\in\bm{\mathcal{S}}$ represents the core semantics. Moreover, in \eqref{eq_sf}, $p(\cdot)$ denotes the pooling function, $\bigoplus$ denotes the cumulative concatenation, and  $\mathrm{Lin}_j[\cdot]$ represents the $j$-th resizing linear module so as to map the concatenated features to match the size of $\mathbf{w}_j$, given the pre-trained generator.  Correspondingly, $\mathbf{s}_j$ is optimised to approach $\mathbf{w}_j$ for our GIE, in which $\bm{\mathcal{S}}=\{\mathbf{s}_1, \ldots, \mathbf{s}_j, \ldots, \mathbf{s}_m\}$ and $\mathbf{f}$ are optimised to reconstruct the image $\mathbf{x}$, given the StyleGAN generator. Both $\mathbf{f}$ and $\bm{\mathcal{S}}$ are thus required to be compressed.} 

\begin{figure}
   \begin{center}
   \includegraphics[width=0.96\linewidth]{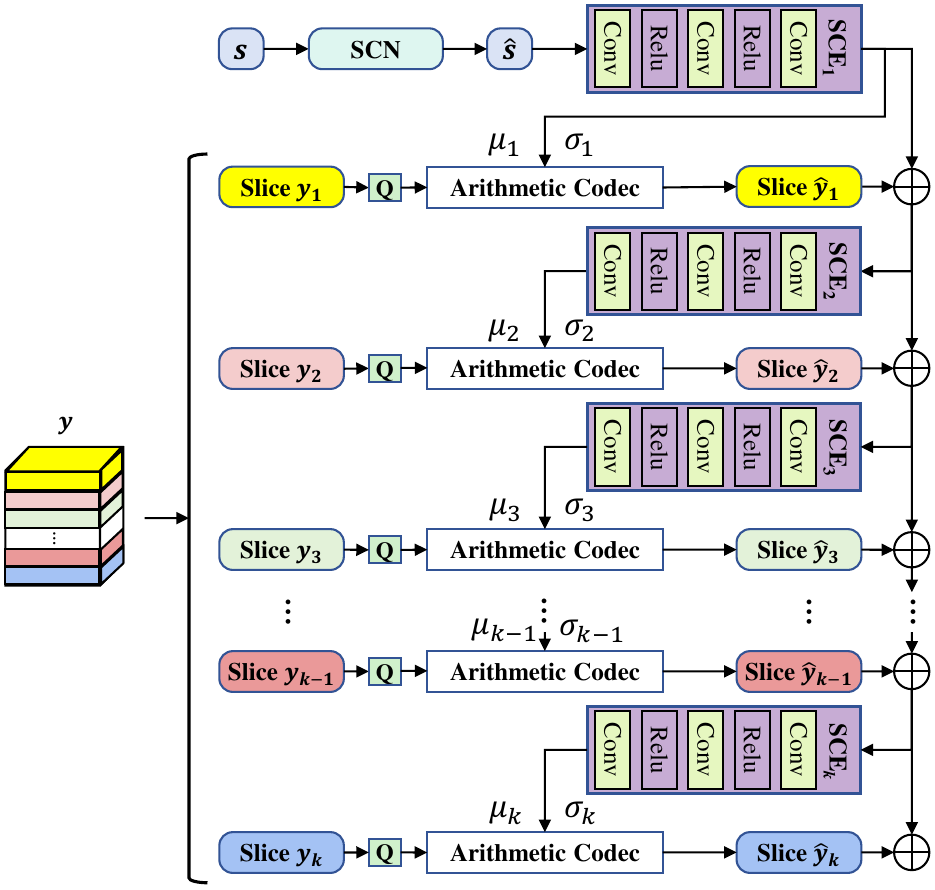}
   \end{center}
   \vspace{-1.5em}
      \caption{{The proposed semantic context model to autoregressively compress semantic features $\bm{y}$ across channel slices $\{\bm{y_1}, \bm{y_2}, \dots, \bm{y_k}\}$, in a coarse to fine-grained manner. Our semantic context encoder network aims to model the pdf of each slice, by predicting the mean $\bm{\mu}$ and diagonal variance $\bm{\sigma}$ given all the previous semantic information. The core semantics from SCN are employed by the first slice as the prior. }}
   \label{fig:channel_entropy}
\end{figure}

{Then, the proposed FCN compresses the middle-level semantic feature $\mathbf{f}$ in the hierarchical style, which includes the feature encoder, semantic context model and feature decoder. The middle-level semantic feature $\mathbf{f}$ is first processed by the feature encoder to reduce the spatial redundancy, which is transformed to a lower-dimensional feature $\bm{y}$ and then divided into $k$ slices $\{\bm{y_1}, \bm{y_2}, \dots, \bm{y_k}\}$. The slices are then compressed by our semantic context model, based on the channel-wise autoregressive entropy models \cite{minnen2020channel} as illustrated in Fig. \ref{fig:channel_entropy}. More specifically, we develop the semantic context encoder to model each compressed slice $\bm{\hat{y}}_i$ as a Gaussian distribution with mean $\bm{\mu}_i$ and diagonal variance $\bm{\sigma}_i$, by aggregating all the previous semantics, including the core semantic cues provided by our SCN when compressing $\bm{\mathcal{S}}$. We shall shortly introduce our SCN in the next Section \ref{sec:method-C}. This can be represented by:
\begin{equation}
[\bm{\mu}_i, \bm{\sigma}_i]=\mathrm{SCE}_i[\mathrm{SCE}_1[\{\mathbf{\hat{s}}_j\}_{j=1}^m]\oplus\bm{\hat{y}}_1\oplus\bm{\hat{y}}_2\oplus...\bm{\hat{y}}_{i-1}]
\end{equation}
where $\mathrm{SCE}_i$ denotes the $i$-th convolutional semantic context encoder and $\oplus$ denotes concatenation across channels. Note that the first slice $\bm{\hat{y}}_1$ is conditioned on the compressed core semantics $\bm{{\hat{\mathcal{S}}}}=\{\mathbf{\hat{s}}_j\}_{j=1}^m$, where $[\bm{\mu}_1, \bm{\sigma}_1]$ are obtained through $\mathrm{SCE}_1[\{\mathbf{\hat{s}}_j\}_{j=1}^m]$. As the number of decoded slices increases, the size of the prior context and the input channels of the context encoder also increase, predicting the distribution with improved accuracy. Then, the pdf of $\bm{\hat{y}}_i$ is modelled as
\begin{equation}
\label{eq:R_F_p}
p_{\bm{\hat{\mathcal{Y}}_i}}(\bm{\hat{y}_i}\mid{\bm{\hat{y}_{i-1}}, \ldots, \bm{\hat{y}_1}}, \{\mathbf{\hat{s}}_j\}_{j=1}^m) \sim \mathcal{N}(\bm{\mu}_i, \bm{\sigma}^2_i).
\end{equation}
This way,  our FCN efficiently compresses $\mathbf{f}$, by the core semantic cues provided by our SCN when compressing $\bm{\mathcal{S}}$. } 

{Finally, the bitrate of the middle-level semantic feature $\mathbf{f}$ can be estimated by the sum of bitrates from all slices as follows,
\begin{equation}
\label{eq:R_F}
\mathcal{R}_{\bm{\mathcal{F}}}=\sum_{i=1}^k\mathbb{E}{[-\log_2(p_{\bm{\hat{\mathcal{Y}}_i}}(\bm{\hat{y}_i}\mid{\bm{\hat{y}_{i-1}},...\bm{\hat{y}_1}, \{\mathbf{\hat{s}}_j\}_{j=1}^m))]}}
\end{equation}
The models can be interpreted as autoregressive along the channel dimension rather than the spatial dimensions which can improve the encoding speed at low bitrates. We use compressed $\bm{{\hat{\mathcal{S}}}}=\{\mathbf{\hat{s}}_j\}_{j=1}^m$ as the prior for the first slice, consistent with the classical hyperprior structure. Since both $\bm{\mathcal{S}}$ and $\bm{\mathcal{F}}_n$ are extracted via the general inversion encoder, they are inherently correlated and redundant.}

\vspace{-0.5em}

\subsection{Core Semantic Compression Network}
\label{sec:method-C}
{Apart from the middle-level semantic features, the core semantics 
$\bm{\mathcal{S}}=\{\mathbf{s}_j\}_{j=1}^m$ still require to be compressed. We introduce our SCN module that first reduces the redundancy by our semantic encoder, represented by $\mathbf{s}_r$. Given the disentangled characteristics of style codes from StyleGAN, we develop the factorized entropy model $\bm{\psi}$ to compress $\mathbf{s}_r$, which is primarily an autoencoder composed of residual convolutional blocks and nonlinear activations, such that the entropy model is differentiable with useful gradients flow through the quantized latent representation. The probability of $\mathbf{s}_r$ could be formulated as
\begin{equation}
p_{\bm{\hat{\mathcal{S}}}_r}(\mathbf{\hat{s}}_r\mid\bm{\psi})=\prod_i(p_{\bm{\hat{\mathcal{S}}}^i_r}(\bm{\psi})*\mathcal{U}(-\frac12,\frac12))(\mathbf{\hat{s}}_r^i),
\end{equation}
where $\mathbf{s}^i_r$ denotes the $i$-th element of $\mathbf{s}_r$, and the predicted density is convolved with a standard uniform density function $\mathcal{U}$ to ensure sufficient flexibility for the priors and align with the variational posterior. Therefore, the bitrate $\mathcal{R}_{\bm{\mathcal{S}}}$ for compressing the core semantics $\bm{\mathcal{S}}$ can be modelled as:
\begin{equation}
\label{eq:R_S}
\mathcal{R}_{\bm{\mathcal{S}}} = \mathbb{E}{[-\log_2(p_{\bm{\mathcal{\hat{S}}}_r}(\mathbf{\hat{s}}_r\mid\bm{\psi}))]}
\end{equation}}

{More importantly, since $\bm{\mathcal{F}}_n$ is the intermediate feature in StyleGAN that essentially includes partial information from the core semantics $\bm{\mathcal{S}}$, we separate the set of the compressed semantics $\bm{\mathcal{\hat{S}}}$ by the size of compressed $\mathbf{\hat{f}}$, as follows,
\begin{equation}
\label{eq:s_w}
\bm{\mathcal{\hat{S}}} = \{\mathbf{\hat{s}}_1, \mathbf{\hat{s}}_2, ... \mathbf{\hat{s}}_m\} = \{\mathbf{\hat{s}}_{1:t}, \mathbf{\hat{s}}_{t+1:m}\} = \{\bm{\mathbf{\hat{s}}}_s, \bm{\mathbf{\hat{s}}}_l\},
\end{equation}
where $\bm{\mathbf{\hat{s}}}_s$ represents the semantics concatenated to generator networks that reside on early layers with resolutions smaller that the size of $\mathbf{\hat{f}}\in\bm{\mathcal{F}}_n$, $\bm{\mathbf{\hat{s}}}_l$ denotes the remaining set for larger resolutions, and $t$ denotes the splitting factor. Apparently, $\bm{\mathbf{\hat{s}}}_s$ denotes the core semantics for editing, which shall be elaborated in Section \ref{sec:method-E} shortly. In contrast,  $\bm{\mathbf{\hat{s}}}_l$ represents the detailed semantics, which are mainly employed for the reconstruction. As $t$ increases, the size of feature tensor $\mathbf{\hat{f}}$ expands, improving reconstruction accuracy and subjective consistency, at the cost of increased bitrates and reduced semantic editing capability. Thus, $t$ is a crucial hyper-parameter to balance the rate-distortion-editing trade-off, which is closely related to $\alpha$ in (\ref{eq_mix}), and is decided based on the resolution of the compressing images. The generator is correspondingly divided by $\mathbf{\hat{f}}$ into two parts $\mathrm{G}=\{\mathrm{G}_s, \mathrm{G}_l\}$ and the  pre-trained $\mathrm{G}_l$ is used to generate the compressed image $\mathbf{\hat{x}}$ based on $\mathbf{\hat{s}}_l$ and $\mathbf{\hat{f}}$ by 
\begin{equation}
\label{eq:g2}
\mathbf{\hat{x}} = \mathrm{G}_l[\mathbf{\hat{s}}_l, \mathbf{\hat{f}}].
\end{equation}
We may need to point out that our method also supports image reconstruction using only $\mathrm{G}[\{\hat{\mathbf{s}}_j\}_{j=1}^m]$. Although its fidelity for human vision is inferior to the hierarchical way, it performs well in machine vision tasks at extremely low bitrates. }

 % And then, we employ the SCN, comprising a semantic residual encoder and decoder, to compress $\bm{s}$ and share its semantic context as a prior to compress $\bm{f}$ using the FCN, which is based on a feature residual encoder and decoder. This process yields the compressed representations $\bm{\hat{s}} = \{\bm{\hat{s}}_1, \bm{\hat{s}}_2\}$ and $\bm{\hat{f}}$ with a low bitrate. 

%  Thus, the total bitrate $\mathcal{R}$ of the network comprises two components, $\mathcal{R}_{\bm{\mathcal{S}}}$ and $\mathcal{R}_{\bm{\mathcal{F}}_N}$, which are jointly optimized under constraint.
% \begin{equation}
% \label{eq:R_all}
% \mathcal{R}=\mathcal{R}_{\bm{\mathcal{S}}} + \mathcal{R}_{\bm{\mathcal{F}}_N}
% \end{equation}
\vspace{-0.5em}
\subsection{Rate-Distortion Optimization}
\label{sec:method-D}
{The training procedure of our HSC method aims to improve both fidelity and editing capabilities given bit-rates, for improving efficiency of semantic compression. More specifically, our training operates in three stages. First of all, our GIE network is first optimised by the inversion style, to ensure the reconstruction and editing capabilities of the network, without considering the bit-rates for compression. At the second stage, the compression networks, including FCN and SCN, are trained to improve the compression efficiency. Jointly training is finally employed to ensure that the network reaches the optimal performance.}

{\noindent {\bf{GIE Training:}}
For training our GIE network, we employ the mean squared error (MSE) loss for pixel-wise supervision and multi-scale LPIPS loss to constrain perceptual similarities across different scales, enhancing inversion accuracy. Given the original image $\mathbf{x}$ and reconstructed image $\mathbf{\hat{x}}$, We thus formulate the two losses as follows,
\begin{align}
\mathcal{L}^{\mathbf{x}}_\mathrm{mse}&=||\hat{\mathbf{x}} - \mathbf{x}||^2_2 \\
\mathcal{L}^{\mathbf{x}}_{\mathrm{mlpips}}&=\sum_{i=0}^2{|| \mathrm{AN}[\lfloor{\mathbf{\hat{x}}}\rfloor_i] - \mathrm{AN}[\lfloor{\mathbf{\hat{x}}}\rfloor_i]||}
\end{align}
which $\mathrm{AN}[\cdot]$ denotes the AlexNet feature extractor \cite{krizhevsky2012imagenet} and $\lfloor\cdot\rfloor_i$ refers to downsampling by a scale factor $2^{i}$. More importantly, to ensure the semantic consistency of $\mathbf{f}$ with $\{\mathbf{s}_s, \mathbf{s}_l\}$, we constrain $\mathbf{f}$ obtained by GIE to approach $\mathrm{G}_s(\mathbf{s}_s)$. This process ensures that the $\mathbf{f}$ is derived from the semantic space $\bm{\mathcal{S}}$, which ensures the editing capability of our HSC method. Therefore, we define the feature semantic consistency (FSC) loss as follows:
\begin{equation}
\mathcal{L}^{\mathbf{f}}_\mathrm{fsc}=||\mathbf{f} - \mathrm{G}_s[\mathbf{s}_s]||^2_2.
\end{equation}
Then, we train our GIE by:
\begin{equation}
\label{eq: trainloss1}
\mathcal{L}_{\mathrm{GIE}}=\mathcal{L}^{\mathbf{x}}_\mathrm{mse}+\lambda_1\mathcal{L}^{\mathbf{x}}_{\mathrm{mlpips}}+\lambda_2\mathcal{L}^{\mathbf{f}}_\mathrm{fsc}
\end{equation}
Moreover, for the face dataset, we additionally adopt the multi-layer identity loss and the face parsing loss introduced by \cite{wei2022e2style} to further ensure the subjective consistency of our GIE, which can be formulated as follows,
\begin{equation}
\mathcal{L}^{\mathbf{x}}_{\mathrm{face}}=\sum_{i=1}^5[2-\langle{\mathrm{AF}_i(\hat{\mathbf{x}}),\mathrm{AF}_i(\mathbf{x})}\rangle-\langle{\mathrm{FP}_i(\hat{\mathbf{x}}),\mathrm{FP}_i(\mathbf{x})}\rangle]
\end{equation}
where $\langle\cdot\rangle$ is cosine similarity, $\mathrm{AF}[\cdot]$ means pre-trained ArcFace network \cite{deng2019arcface} and $\mathrm{FP}[\cdot]$ is pre-trained face parsing model \cite{CelebAMask-HQ}. }

{\noindent {\bf{FCN and SCN Training:}}
Given the pre-trained GIE and StyleGAN generator, we then train our FCN and SCN according to the middle-level semantic feature $\mathbf{f}$ and the core semantics $\{\mathbf{s}_j\}_{j=1}^m$.  Losses for rate-distortion optimisation are employed as,
\begin{align}
\label{eq:trainloss2}
\mathcal{L}_{\mathrm{SCN}}&=\mathcal{R}_{\bm{\mathcal{S}}} + \lambda_{3}\sum_{j=1}^m||\mathbf{\hat{s}}_j - \mathbf{s}_j||^2_2 \\
\label{eq:trainloss3}
\mathcal{L}_{\mathrm{FCN}}&=\mathcal{R}_{\bm{\mathcal{F}}} + \lambda_{4}||\mathbf{\hat{f}} - \mathbf{f}||^2_2
\end{align}
where $\mathcal{R}_{\bm{\mathcal{S}}}$ and $\mathcal{R}_{\bm{\mathcal{F}}}$ are the bit-rates calculated by (\ref{eq:R_S}) and (\ref{eq:R_F}). Due to the intrinsic semantic reflected in $\mathbf{f}$ and  $\{\mathbf{s}_j\}_{j=1}^m$, we found that at this second stage, without considering the consistency and perceptual losses at the image level and only training on $\mathbf{f}$ and  $\{\mathbf{s}_j\}_{j=1}^m$ can achieve fast convergence.}

{
\noindent {\bf{Jointly Training:}}
Building upon the warm-up training from the previous two stages,  we perform the joint training of our HSC method, with the following losses to optimize the entire framework:
\begin{equation}
\label{eq:trainloss4}
\mathcal{L}=\mathcal{L}_{\mathrm{SCN}}+\mathcal{L}_{\mathrm{FCN}}+\lambda_5\mathcal{L}_{\mathrm{GIE}}
\end{equation}}

\subsection{Semantic Image Editing}
\label{sec:method-E}
\begin{figure}[htbp]
   \begin{center}
   \includegraphics[width=1\linewidth]{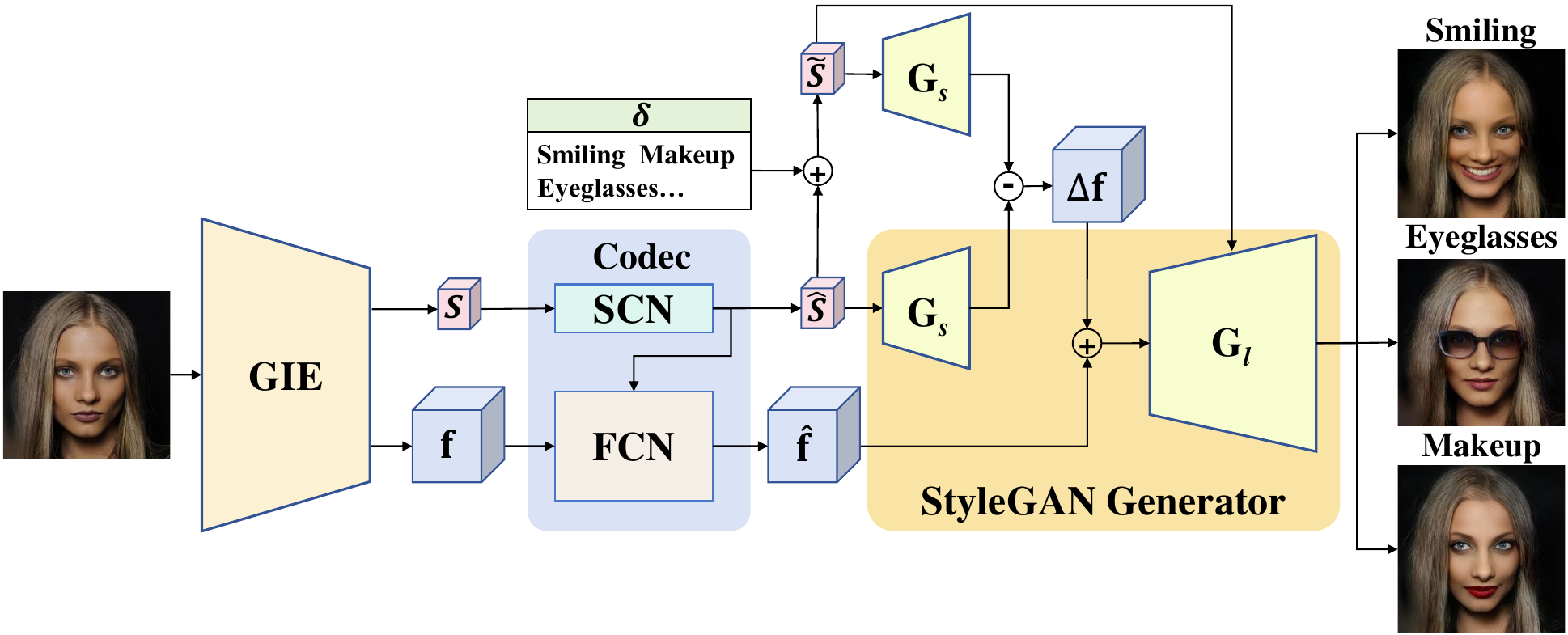}
   \end{center}
   \vspace{-1em}
      \caption{{The proposed HSC method to edit the compressed images by our compressed middle-level semantic feature $\mathbf{\hat{f}}$ and and core semantics $\bm{\mathcal{\hat{S}}}$. We illustrate the editing by smiling, eyeglasses and makeup as examples, which achieve accurate and realistic editing.}}
   \label{fig:editpipeline}
   \vspace{-1em}
\end{figure}

{A core characteristic of our HSC method is the ability of retraining semantics, which is able to achieve semantic consistent compression as previously shown in Fig. \ref{fig:sc} and also realistic editing. Given the disentangled directions $\bm{\delta}\in\bm{\mathcal{S}}$ given pre-trained StyleGAN for certain attributes \cite{harkonen2020ganspace}, we perform image editing based on the compressed $\bm{\hat{\mathcal{S}}}$ and $\hat{\mathbf{f}}$, during the inference stage \cite{yao2022style}, which is illustrated in Fig. \ref{fig:editpipeline}. }

\begin{figure*}[htbp]
   \begin{center}
   \includegraphics[width=1\linewidth]{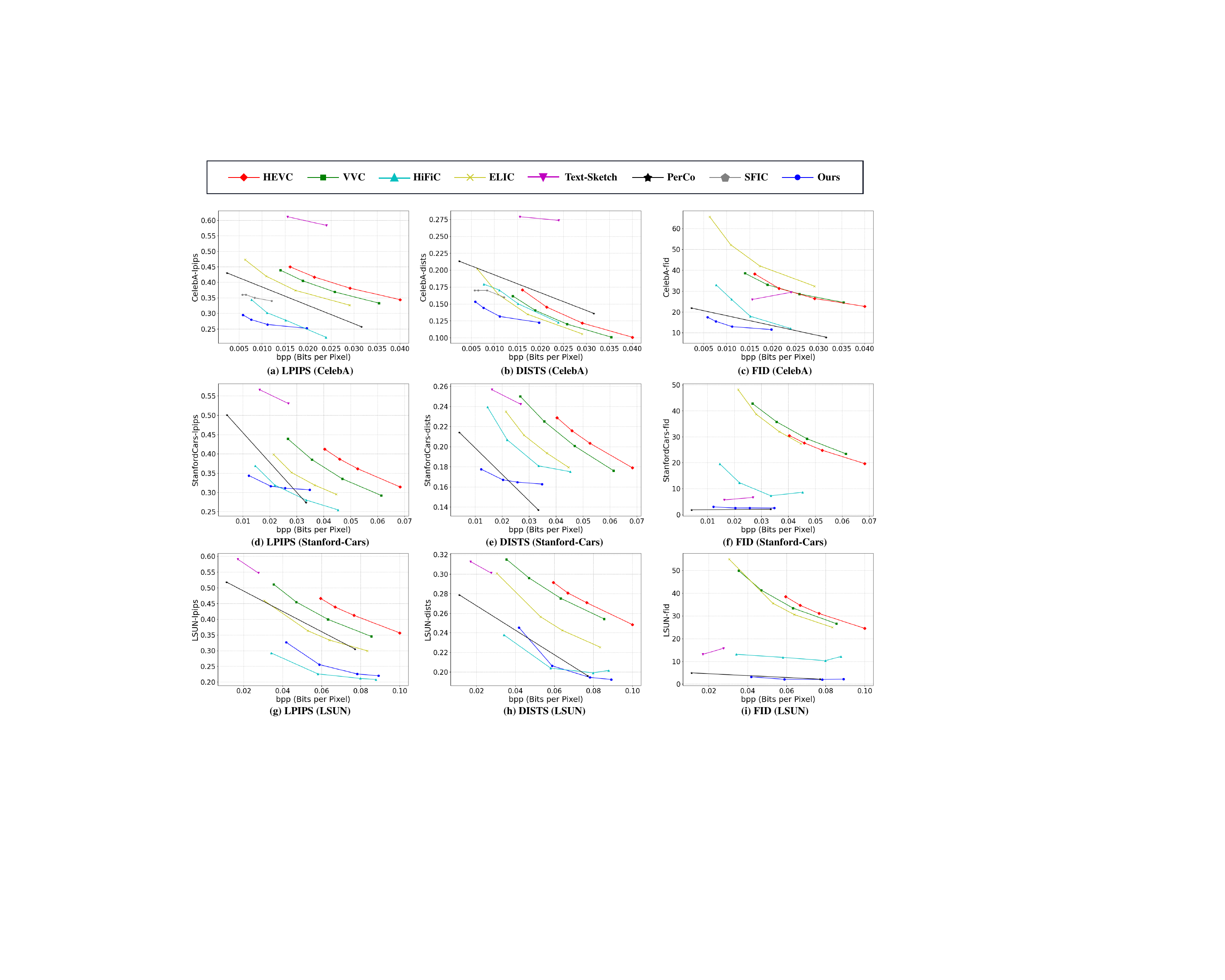}
   \end{center}
   \vspace{-1em}
      \caption{{Rate-distortion curves of HEVC \cite{sullivan2012overview}, VVC \cite{bross2021overview},  HiFiC \cite{mentzer2020high}, ELIC \cite{he2022elic}, Text-Sketch \cite{lei2023text+sketch}, PerCo \cite{careil2023towards}, SFIC \cite{mao2023scalable} and our HSC methods, in which the distortion is measured by LPIPS, DISTS and FID metrics on 3 datasets. }}
   \label{fig:curve_lpips}
   \vspace{-1em}
\end{figure*}

{More specifically, given the editing direction $\bm{\delta}$, we obtain the new core semantics as 
\begin{equation}
\bm{\tilde{\mathcal{S}}} = \{\mathbf{\tilde{s}}_s, \mathbf{\tilde{s}}_l\} = \{\mathbf{\hat{s}}_s+\bm{\delta}, \mathbf{\hat{s}}_l+\bm{\delta}\}.
\end{equation}
Unlike other GAN inversion methods that edit only core semantic latent codes, our HSC method hierarchically includes the middle-level semantic feature $\mathbf{f}$. We thus need to calculate the corresponding editing direction for $\mathbf{f}$ to achieve the desired editing on attributes and images; this is calculated by subtracting the features generated from the compressed $\mathbf{\hat{s}}_s$ and the interpolated $\mathbf{\tilde{s}}_s$ as follows,
\begin{equation}
\Delta{\mathbf{f}} = \mathrm{G}_s[\mathbf{\tilde{s}}_s] - \mathrm{G}_s[\mathbf{\hat{s}}_s].
\end{equation}
Recall that $\mathrm{G}_s[\mathbf{\hat{s}}_{s}]$ represents the feature from the original semantics, while $\mathrm{G}_s[\mathbf{\tilde{s}}_{s}]$ represents the feature reconstructed by the editing semantics. Given the middle-level semantics within $\mathbf{f}$, $\Delta{\mathbf{f}}$ thus stands for the deviation provided by $\bm{\delta}$ on $\mathbf{\hat{f}}$. Furthermore, given $\mathrm{G}_l$ as the remaining part of the generator, we can restored the editing image as follows,
\begin{equation}
\mathbf{\tilde{x}} = \mathrm{G}_l[\mathbf{\tilde{s}}_l,\mathbf{\tilde{f}}]=\mathrm{G}_l[\mathbf{\hat{s}}_l+\bm{\delta}, \mathbf{\hat{f}} + \Delta\mathbf{{f}}].
\end{equation}}

{Therefore, given the fact that existing compression methods, even for existing semantic compression methods, primarily focus on achieving low bitrate while maintaining high reconstruction quality. Our HSC method, besides achieving the state-of-the-art rate-distortion performances, also is the first work capable of achieving semantic editing and subjective consistency along with increasing bit-rates. This essentially coincides with human visual system in compressing images, thus providing a new framework for future image/video compression paradigms.}

\section{Experiment}

%To assess the performance of the proposed semantic compression framework, we conducted experiments from two perspectives: compression efficiency and semantic characteristics. For compression efficiency, we compared our framework against a range of existing methods, including traditional codecs, learning-based approaches, generative compression, and semantic compression techniques. To evaluate semantic characteristics, we performed three distinct experiments: semantic editing, semantic segmentation, and semantic style mixing to comprehensively validate the semantic capabilities of the model.

\subsection{Experimental Settings}

{\noindent {\bf{Datasets:}} We comprehensively evaluated our HSC method based on various scenarios, including the face, car and outdoor church test images. More specifically, the FFHQ dataset,  consisting of  70,000 facial images of $1024\times1024$ resolutions \cite{karras2019style}, was employed for the face scenario, together with the Stanford-Cars \cite{krause20133d} and LSUN \cite{yu15lsun} datasets during the training stage.  For the inference, we evaluated the compression and editing performances based on the CelebA-HQ dataset \cite{Karras2017ProgressiveGO}, in which 5,000 images of $1024\times 1024$ resolution were randomly selected from 30,000 images. For the car and church domains, we performed the evaluation on the corresponding test sets, each of which comprises 300 images of $384\times 512$ and $256\times 256$ resolutions. Furthermore, we conducted the assessment on face parsing task upon the CelebA-Mask-HQ test dataset and pre-trained face parsing framework \cite{CelebAMask-HQ}, so as to validate the performance on the emerging video coding for machine (VCM) scenario.}
\begin{figure*}[htbp]
   \begin{center}
   \includegraphics[width=0.975\linewidth]{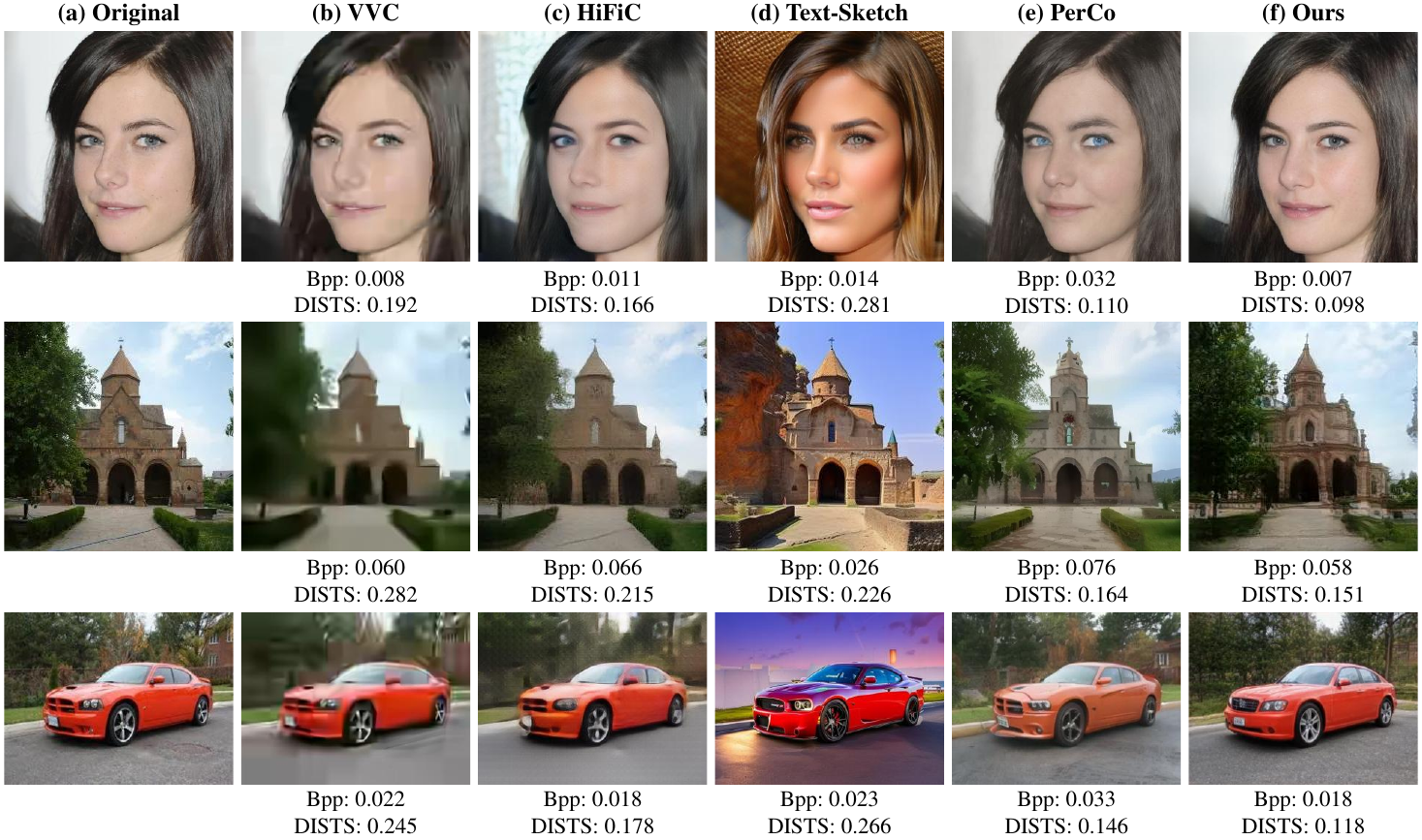}
   \end{center}
   \vspace{-1.5em}
      \caption{{Subjective results for representative compression methods under similar bit-rates, for the face, outdoor church and car scenarios. We also report the bpp and DISTS metric to quantify compression efficiency.}}
   \label{fig:subjective}
   \vspace{-1em}
\end{figure*}

\begin{figure}[htb]
   \begin{center}
   \vspace{-0.5em}
   \includegraphics[width=0.95\linewidth]{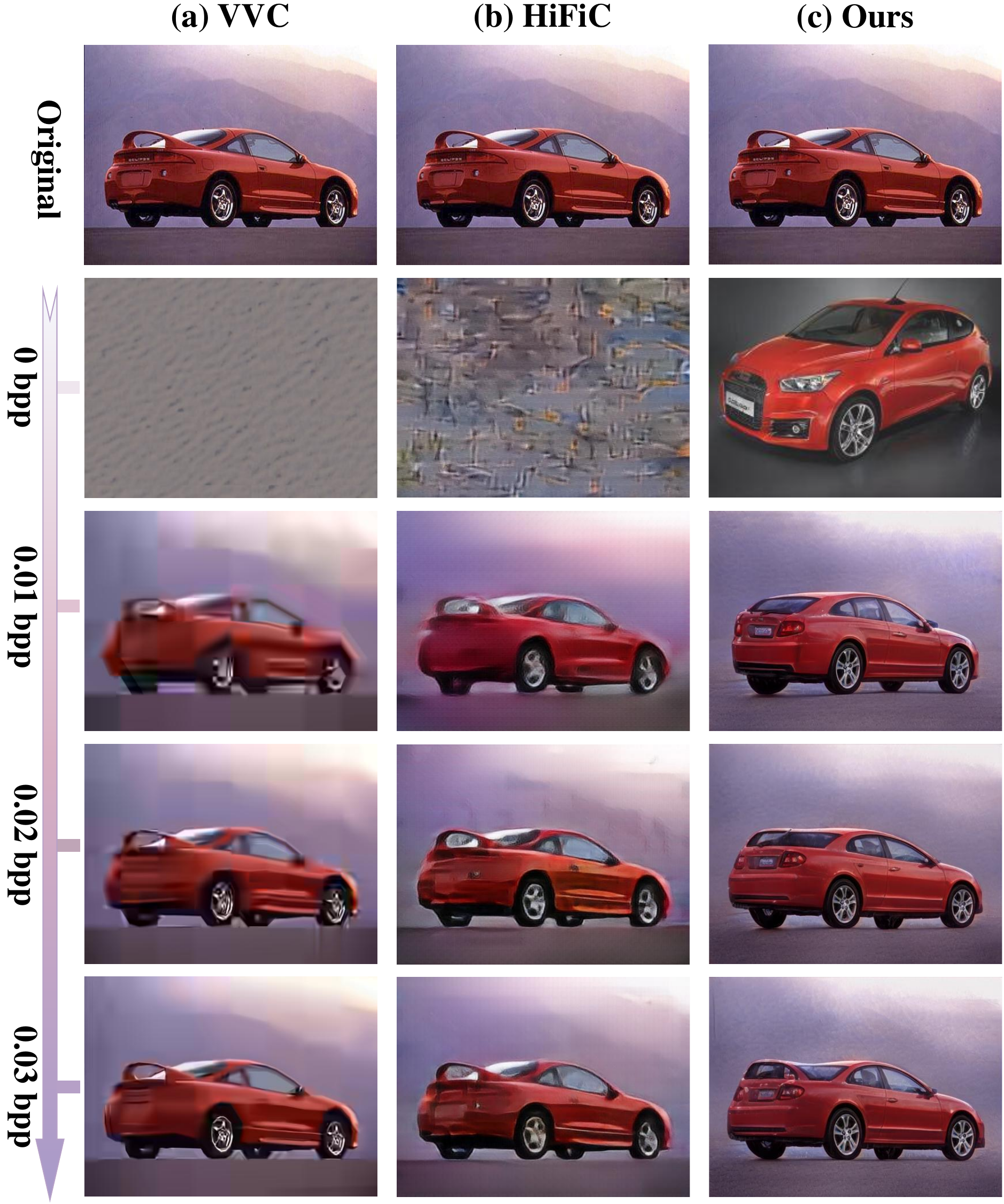}
   \end{center}
   \vspace{-1em}
      \caption{Subjective results of representative compression methods by varying bit-rates when compressing car images, which include VVC for the state-of-the-art standard codec, HiFiC for the subjective-oriented compression, and our HSC for semantic compression. Please note that our HSC method achieves semantic consistency when increasing the bit-rates from 0 bpp.}
   \label{fig:bigsub}
   \vspace{-1.5em}
\end{figure}
{\noindent {\bf{Metrics:}} To evaluate the compression efficiency, we measured the perceptual distortion by the widely used reconstruction metrics, including the learned perceptual image patch similarity (LPIPS) \cite{zhang2018unreasonable}, deep image structure and texture similarity (DISTS) \cite{ding2020image}, and Fréchet Inception distance (FID) \cite{heusel2017gans}. Lower LPIPS and DISTS scores indicate better visual fidelity, while a lower FID score also reflects better alignment with real images, thus preserving semantic consistency. More importantly, due to limited numbers of images in the Stanford-Cars and LSUN test datasets, we used 192-dimensional features from the \textit{de facto} Inception\_V3 model to calculate the FIDs, whereas for the CelebA dataset, the default 2048 dimensions were kept. Furthermore, we evaluated semantic segmentation performance using frequency-weighted intersection over union (FwIoU) \cite{garcia2017review} to assess the capabilities of retaining semantics for VCM.}

\noindent {\bf{Implementation Details:}} In our HSC method, we set $t=5$ for high-resolution face images, such that the spatial size for $\mathbf{f}$ was $16\times16$. For the relatively low resolutions, we set $t=3$, resulting in $\mathbf{f}$ of $8\times8$. Moreover, $k=8$ in \eqref{eq:R_F} was empirically selected for semantic context model to balance the performance and efficiency. 
%For the training, we trained the GIE at the first stage for 15 epochs according to \eqref{eq: trainloss1}. At the second stage, we trained our FCN and SCN for 10 epochs respectively according to \eqref{eq:trainloss2} and \eqref{eq:trainloss3}. For the final stage, we jointly trained our HSC method for 15 epochs according to \eqref{eq:trainloss4}. 
For all the training stages,  the base learning rate was set to 0.0001 and we used the Ranger optimizer, which is a combination of Rectified Adam \cite{liu2019variance} with the Lookahead \cite{zhang2019lookahead} technique, with exponential decay rates ($\beta_1$,$\beta_2$) = (0.95, 0.999). The overall training was based on one NVIDIA 4090 GPU for approximately two days. Our code for both training and inference shall be released upon acceptance.

\begin{table*}[htbp]
\centering
\caption{BD related results based on the VVC codec on the CelebA, Stanford-Cars, and LSUN datasets, which essentially depict the improvements on the metrics given the same bit-rates. A downward arrow ($\downarrow$) indicates that a lower score is better for restoration accuracy and consistency. Red font highlights the best values, while blue indicates the second-best values.}
\label{tab:table1}
\resizebox{0.8\textwidth}{!}{
\begin{tabular}{cc|ccc|ccc|ccc}
\hline
\hline
\multicolumn{2}{c|}{\textbf{Datasets}}  &\multicolumn{3}{c|}{{\textbf{CelebA}}} &\multicolumn{3}{c|}{\textbf{Stanford-Cars}} & \multicolumn{3}{c}{\textbf{LSUN}}\\ \hline
\multicolumn{2}{c|}{Methods}& {BD-L$\downarrow$} & {BD-D$\downarrow$} & {BD-F$\downarrow$} & {BD-L$\downarrow$} & {BD-D$\downarrow$} & {BD-F$\downarrow$} & {BD-L$\downarrow$} & {BD-D$\downarrow$} & {BD-F$\downarrow$}\\\hline
\multicolumn{2}{c|}{VVC \cite{bross2021overview}}    &0.000 &0.000 &0.00 &0.000 &0.000 &0.00 &0.000 &0.000 &0.00         \\
\multicolumn{2}{c|}{HEVC \cite{sullivan2012overview}}  &0.026  &0.012  &0.14 &0.046 &0.013 &-1.97 &0.049 &0.009 &2.43        \\
\multicolumn{2}{c|}{ELIC \cite{he2022elic}}  &-0.038 &-0.011 &6.79 &-0.065 &-0.030 &-3.06 &-0.067 &-0.031 &-2.18  \\
\multicolumn{2}{c|}{SFIC \cite{mao2023scalable}}  &-0.105 &\color{blue}{-0.012}  & - & - & - & - & - & - & - \\
\multicolumn{2}{c|}{HiFiC \cite{mentzer2020high}} &\color{blue}{-0.150} &-0.005 &\color{blue}{-19.00} &\color{red}{\textbf{-0.113}} &\color{blue}{-0.046} &\color{blue}{-28.62} &\color{red}{\textbf{-0.190}} &\color{red}{\textbf{-0.075}} &\color{blue}{-24.77}    \\
\multicolumn{2}{c|}{Ours}   &\color{red}{\textbf{-0.165}} &\color{red}{\textbf{-0.026}} &\color{red}{\textbf{-23.66}} &\color{blue}{-0.106} &\color{red}{\textbf{-0.075}} &\color{red}{\textbf{-36.98}} &\color{blue}{-0.150} &\color{blue}{-0.069} &\color{red}{\textbf{-32.83}} \\ \hline
\multicolumn{11}{l}{Note: BD-L denotes the BD-LPIPS, BD-D is BD-DISTS and  BD-D is BD-FID to represent the change in }\\
\multicolumn{11}{l}{the  metric between different compression methods at the same bitrate.}\\
\hline\hline
\end{tabular}
}
\vspace{-1em}
\end{table*}

\subsection{Evaluations on Compression Quality}
\label{sec:experiment-B}

{
\noindent {\bf{Baselines:}} Regarding semantic compression, we compared the proposed HSC method with the state-of-the-art compression standards and advanced learning-based methods. 
\begin{itemize}
\item{\bf{VVC}}: We employed the all-intra mode in VVC reference software VTM 12.0 and compared the reconstruction results by setting the quantization parameters (QP) to $\{\mathrm{QP}_i\}^4_{i=1}\! =\! \{43, 45, 47, 49\}$ for CelebA dataset and $\{\mathrm{QP}_i\}^4_{i=1} \!=\! \{45, 47, 49, 51\}$ for LSNU and Stanford-Cars datasets.
\item{\bf{HEVC}}: We also employed the all-intra mode in HEVC reference software HM 18.0, with the same QPs as the VVC codec to achieve similar bitrates.
\item{\bf{HiFiC}}: The generative assisted compression method HiFiC \cite{mentzer2020high} was fine-tuned given $\lambda=\{2^4, 2^5, 2^6, 2^7\}$, based on the pre-trained model. For fair comparisons, we trained HiFiC by the same datasets as ours.
\item{\bf{ELIC}}: The state-of-the-art learning-based compression model ELIC \cite{he2022elic} was fine-tuned by $\lambda=\{1, 2.5, 5, 10\}\times10^{-4}$ based on the low-bit pre-trained model and the same datasets as ours.
\item{\bf{Text-Sketch}}: The text-to-image semantic compression method \cite{lei2023text+sketch} was also employed for comparison, given pre-trained models available by $\lambda=\{0.5, 1.0\}$.
\item{\bf{PerCo}}: The state-of-the-art semantic compression method \cite{careil2023towards} based on diffusion models was also compared by the publicly available version \cite{korber2024perco} given $\lambda=\{0.0019, 0.0313\}$.
\item{\bf{SFIC}}: When comparing face image compression, we also compared with the most recent StyleGAN-based compression method SFIC \cite{mao2023scalable}\footnote{Please note that SFIC was particularly applied for face image compression, and we thus compared our HSC method with SFIC only on face images. We also performed comparisons on segmentation parsing tasks.}. 
\end{itemize}
}

{
\noindent {\bf{Quantitative Results:}} We first evaluate the semantic compression performance regarding restoration accuracy, and plot the rate-distortion curves in Fig. \ref{fig:curve_lpips}. As can be seen from this figure, our method almost outperforms all the baselines across all the metrics, especially for low bit-rate scenarios. In particular, due to the fact that StyleGAN benefits from generating extremely realistic images from merely small dimensions, our HSC method surpasses the state-of-the-art diffusion based method, namely, PerCo, enabling our method to be particularly advantageous especially in ultra-low bitrate scenarios. We further report the Bjøntegaard-Delta (BD) related results in Table \ref{tab:table1}, such that the overall fidelity improvements are calculated given the same bit-rates. As can be summarised from this table, our method achieves the best results in almost all metrics, especially for FID. Although HiFiC achieved superior performances in few scenarios, it does not belong to semantic compression, which prohibits it from applications to semantic-related tasks and benefits, as shall be explained in sequel. More importantly, the superiority of our HSC method is further highlighted in compressing high-resolution images, e.g., the $1024\times1024$ resolution for face images.}

{
\noindent {\bf{Qualitative Results:}} We also illustrate qualitative results in Fig. \ref{fig:subjective}.  From this figure, the images compressed by VVC and HEVC codecs exhibit blocking artefacts in low bit-rate scenarios due to the inherent limitations of block-based hybird framework. HiFiC tends to exhibit pronounced checkerboard artefacts in background regions at extremely low bitrates, while our method exhibits natural appearance and accurate details. Moreover, the semantic compression method, namely, Text-Sketch method, generates low-fidelity reconstructions that heavily deviates from the original image in colour and details. {On the other hand, the results of semantic diffusion method PerCo exhibit degradation in visual detail quality, including changes in colour, contour and details}. Our HSC method, consuming similar and even fewer bitrates, consistently achieves the best subjective results with negligent artefacts, accurate details, and realistic content.}

{To further verify the distinct characteristic of our HSC method for semantic compression, we further illustrate the compressed images by varying bitrates in Fig. \ref{fig:bigsub} for car images, apart from the face images in Fig. \ref{fig:sc}. From this figure, we can conclude that our HSC method approaches the input image by semantic consistency, instead of the pixel-wise accuracy from most of the representative baselines. In other words, with the increase of bit-rates, our method gradually restores the semantics from the overall appearances, to the detailed orientation and types of the provided car image. This, on the one hand, enables the superior subjective quality and accurate fidelity of our HSC method even at extremely low bit-rates. On the other hand, the subjective consistency further verifies that our HSC method essentially compresses the intrinsic semantics, which is in accordance to human visual system and is also pronounced in the emerging broad applications for VCM, as shall be elaborated shortly. }

\begin{figure}[htbp]
\centering
   \begin{center}
     \includegraphics[width=0.9\linewidth]{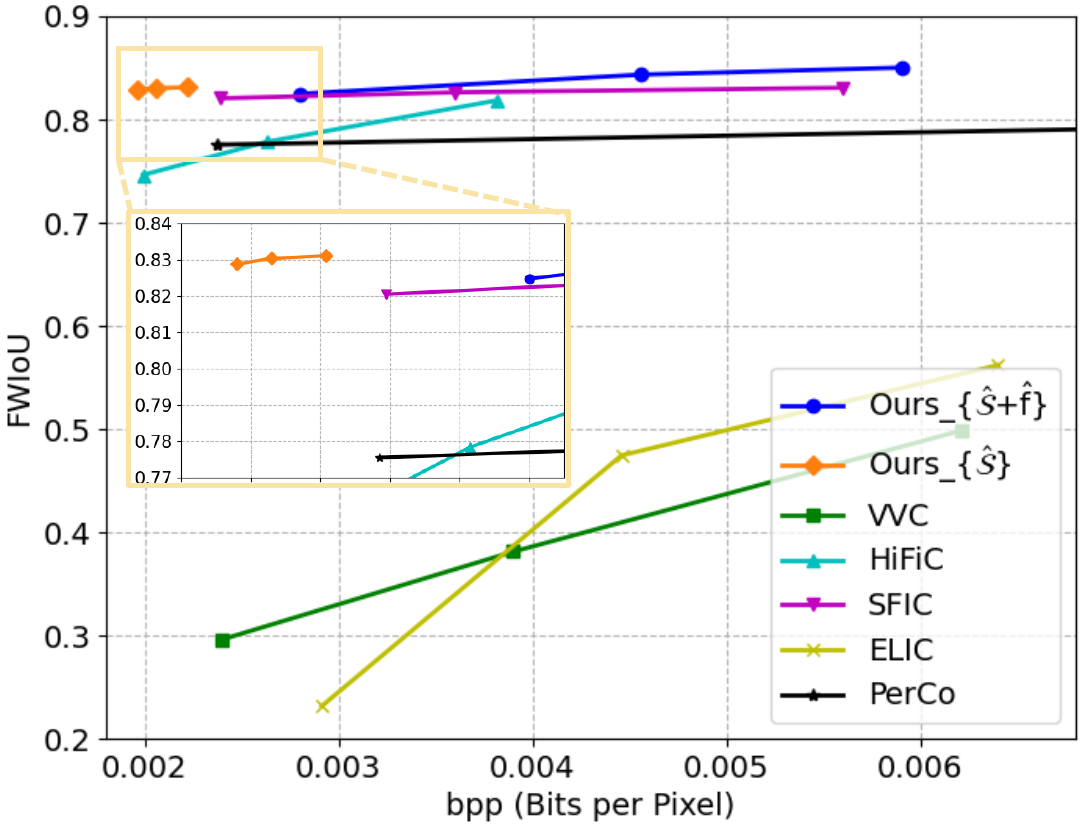}
   \end{center}
   \vspace{-1em}
      \caption{The rate-accuracy curves of VVC, HiFiC, SFIC, ELIC, PerCo and two variants of our HSC method, regarding the machine vision task for face segmentation on the CelebA-Mask-HQ dataset.}
   \label{fig:vcm_line}
   \vspace{-1em}
\end{figure}

% \begin{figure}[ht]
% \centering
% \subfigure[Compressed Smiling Editing]{\includegraphics[width = 0.8\linewidth\centering]{figs/fig7-edit-fid-sub1.pdf}}\vspace{-.5em}
% \subfigure[Compressed Eyeglasses Editing]{\includegraphics[width = 0.8\linewidth]{figs/fig7-edit-fid-sub2.pdf}}\vspace{-.5em}
% \caption{The qualitative comparisons of the proposed HSC method for (a) compressed-smiling and (b) compressed-eyeglasses.}
% \label{fig:edit_fid}
% \vspace{-1em}
% \end{figure}
\begin{figure*}[hbp]
   \begin{center}
   \includegraphics[width=0.975\linewidth]{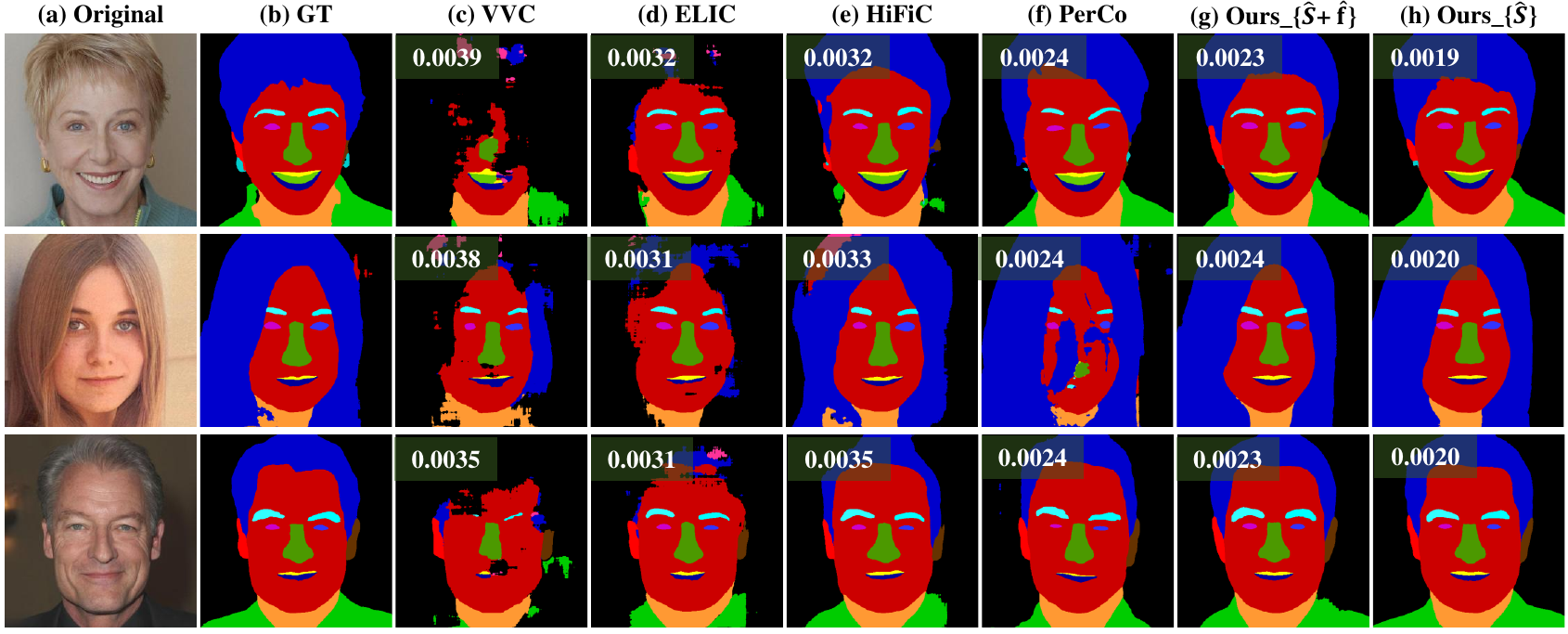}
   \end{center}
   \vspace{-1.5em}
      \caption{Qualitative comparisons for the face recongnition task, based on the compressed images. (b) is the ground truth, (c-f) represent the competitive methods and (g-h) are two variants of our HSC method. (g) indicates reconstructing the image using both $\mathbf{\hat{f}}$ and $\bm{\mathcal{\hat{S}}}$, while (h) represents reconstructing using only $\bm{\mathcal{\hat{S}}}$. The bpp for each image are reported on the corresponding image. We may also need to point out that since the source code of SFIC was not available, we only compared quantitative results, without the comparisons on qualitative results. }
   \label{fig:vcm_subjuctive}
   % \vspace{-0.5em}
\end{figure*}

\subsection{Evaluations on Machine Vision Tasks}
{Another benefit of our HSC method lies in the VCM task, in which we evaluated the baseline methods based on the semantic segmentation for face images. More specifically, we evaluated two variants of our HSC method, namely, using only $\mathrm{G}[\bm{\mathcal{\hat{S}}}]$ to reconstruct images (denoted as $\text{Ours}\_\{\bm{\hat{\mathcal{S}}\}}$) and the default hierarchical framework for both $\bm{\mathcal{\hat{S}}}$ and $\mathbf{\hat{f}}$ (denoted as $\text{Ours}\_\{\bm{\hat{\mathcal{S}}}+\hat{\mathbf{f}}\}$). Regarding the face segmentation task, we illustrate the quantitative results in Fig. \ref{fig:vcm_line}, together with the qualitative results in Fig. \ref{fig:vcm_subjuctive}. From those figures, it is obvious that both two variants of our HSC method achieve superior performance for VCM, i.e., the best FWIoU values given the same bit-rates. Moreover, when solely using the core semantics, $\text{Ours}\_\{\bm{\hat{\mathcal{S}}\}}$ obtains  exceptional performance at extremely low bitrates. With the assistance from the middle-semantic feature $\mathbf{f}$, $\text{Ours}\_\{\bm{\hat{\mathcal{S}}}+\hat{\mathbf{f}}\}$ achieves the best segmentation accuracy. On the other hand,  when the bpp was lower than 0.006, neither ELIC nor VVC methods achieved satisfactory segmentation results. While HiFiC, SFIC and PerCo performed relatively well, there was still a noticeable gap compared to our HSC approach. Therefore, benefiting from our hierarchical architecture in compressing the intrinsic semantics, our HSC method is able to maximally retain the semantic consistency, thus acquiring the state-of-the-art performances in VCM tasks.}

%\li{The qualitative results are shown in Fig.\ref{fig:vcm_subjuctive} and it can be observed that our two methods achieve complete segmentation of all facial features using the least number of bits. ELIC and VVC methods fail to segment the full contour of the face. While HiFiC does manage to segment the face, the edges are relatively rough, indicating a higher degree of distortion in the image's edge details. Since SFIC's source code is not available, we only have quantitative results reported in the paper and no qualitative evaluation. }

\vspace{-0.5em}
\subsection{Altering Compressing Semantics}

\begin{figure*}[htb]
   \begin{center}
   \includegraphics[width=1\linewidth]{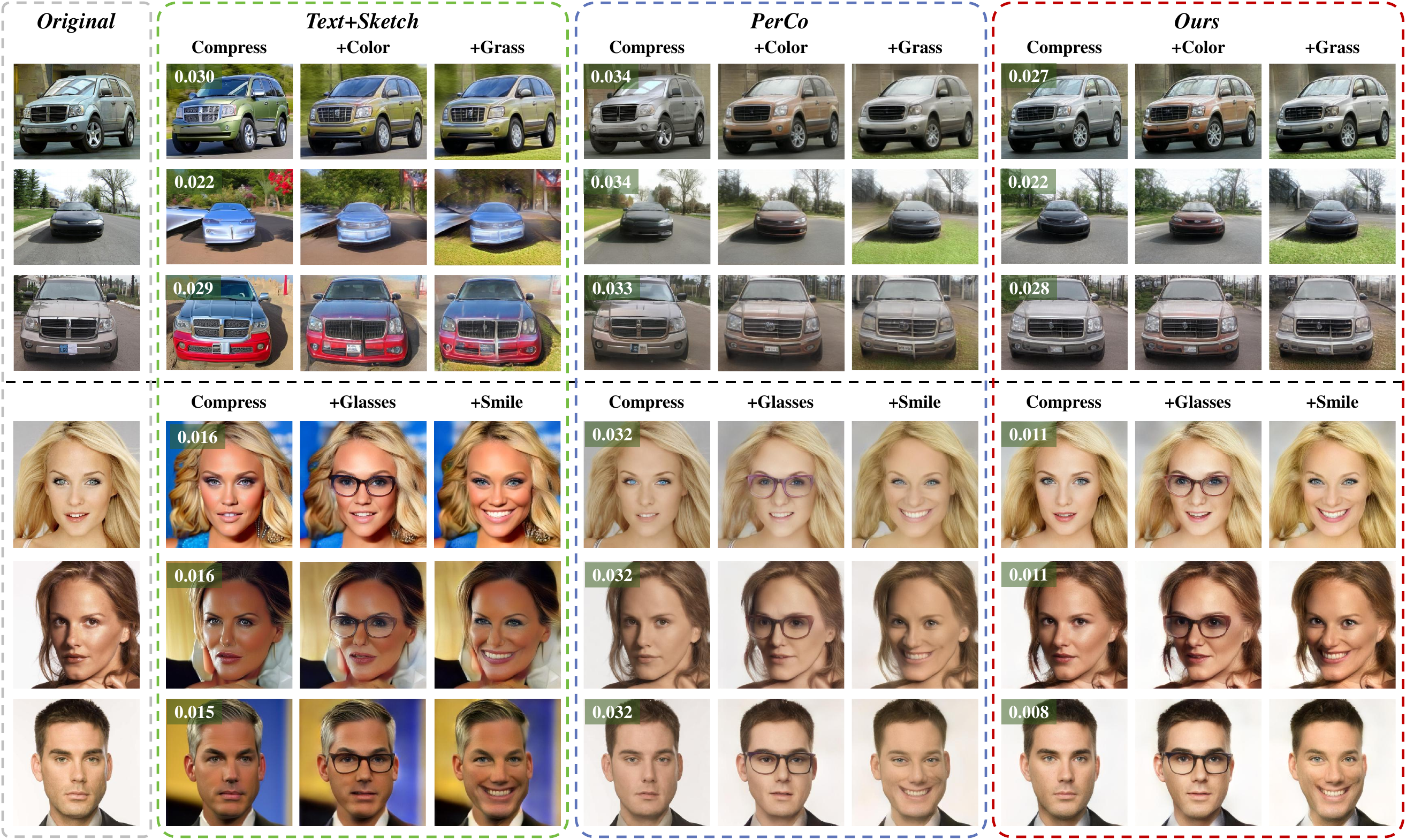}
   \end{center}
   \vspace{-1.5em}
      \caption{Subjective results of image editing based on compressed bitsteams, compared among existing state-of-the-art semantic compression methods, including Text-Sketch, PerCo and our HSC methods. We sequentially exhibit the unaltered compressed images and the editing images by meaningful directions. Please note that we do not report the outdoor church images since to the best of our knowledge, there exist no valid and meaningful editing directions in literature.}
   \label{fig:edit_subjective}
\end{figure*}

{
\noindent {\bf{Semantic Editing:}} Another crucial characteristics of our HSC method is the intrinsic semantics within the compressed bitstreams.  More specifically, we performed the editing operation based on the compressed semantics from our HSC method. The editing directions were obtained by the InterfaceGAN \cite{shen2020interpreting} method for face images, whilst for car images, we employed the GANspace \cite{harkonen2020ganspace} method to find meaningful editing directions. We report the results in  Fig. \ref{fig:edit_subjective}. From this figure, we can find that the Text-Sketch method even can not alter the semantics based on the compressed bitstreams. In contrast, both the PerCo and our HSC methods essentially possess the semantics in the bitstreams. However, the PerCo method, due to the unstructured latent space of diffusion models, falls short in retaining the core identity and details during editing. Our HSC method, with the lowest bitrates, achieves the state-of-the-art performances in both semantic consistency and editing realism.}

\begin{figure}[htbp]
   \begin{center}
   \includegraphics[width=0.96\linewidth]{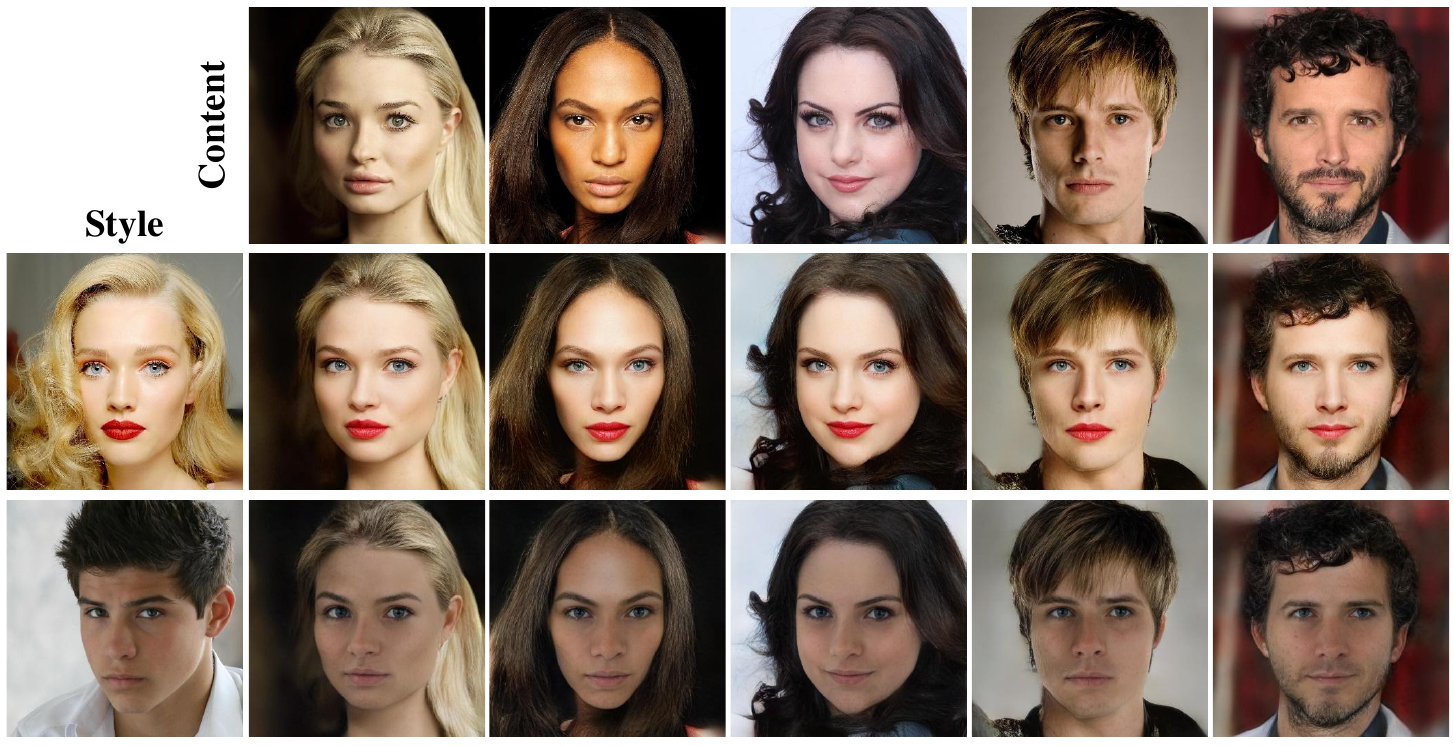}
   \end{center}
   \vspace{-1.5em}
      \caption{Subjective results for mixing style from two images compressed by our HSC method. Each style image corresponds to five content images, in which the compressed core semantics $\bm{\mathcal{\hat{S}}}$ of style images were mixed with the compressed middle-level semantic feature $\mathbf{\hat{f}}$ from content images.}
   \label{fig:mixtype}
   \vspace{-1.5em}
\end{figure}

\noindent {\bf{Style Mixing:}} Given the fact that our HSC method compresses the core semantics and middle-level semantics in a hierarchical way, we also conducted  the style mixing task in Fig. \ref{fig:mixtype}, whereby the mixing was performed by combining the core semantics $\bm{\mathcal{\hat{S}}}$ from style images with the middle-level semantic feature $\mathbf{\hat{f}}$ from content images. From Fig. \ref{fig:mixtype}, it is evident that the structural details, such as pose and facial shape, are encoded by the middle-level semantics $\mathbf{\hat{f}}$, whereas the styles, including eyes colour and makeup, are encoded by the semantic $\bm{\mathcal{\hat{S}}}$. Despite the degradation brought by the compression, the intrinsic semantics by our HSC method still manage to maintain an effective and visually coherent mixing effect on restoring the images.

% \begin{table}[ht]   %  !htbp
% \vspace{-.5em}
% \renewcommand\arraystretch{0.9}
% \begin{center}
%     \small
%     \caption{Facial latent space editing comparisons on the CelebA.}
%     \vspace{.3em}
%     \label{edit_table}
%     \resizebox{0.48\textwidth}{!}{
%     \begin{tabular}{c|c|c|c|c}
%         \hline
%         \hline 
%         \rowcolor{mygray}
%         Method & bpp & Smiling FID & Glasses FID & Make-up FID \\
%         \hline
%         VVC& 72.15 & 1759.15 & 73.99  & 485.63\\\rowcolor{mygray}
%         HEVC& 71.20& 897.33 & 71.81 & 279.49\\
%         HiFiC& 67.73& 456.79 & 69.71 & 170.01\\\rowcolor{mygray}
%         ELIC& 67.73& 456.79 & 69.71 & 170.01\\
%         Text-Sketch& 60.54& 225.11 & 65.26 & 109.35\\\rowcolor{mygray}
%         Ours& 60.54& 225.11 & 65.26 & 109.35\\
%         \hline\hline
%     \end{tabular}
%     }
% \end{center}
% \vspace{-2.5em}
% \end{table}

% \begin{figure*}[htbp]
%    \begin{center}
%    \includegraphics[width=0.96\linewidth]{figs/fig7-edit-fid.pdf}
%    \end{center}
%    \vspace{-1.5em}
%       \caption{The qualitative editing evaluation of the proposed CSCN method.}
%    \label{fig:edit_fid}
%    \vspace{-0.5em}
% \end{figure*}

\subsection{Ablation Study}

Since our HSC method operates in a hierarchical style, we conducted ablation study on the core components, namely, the SCN, FCN and hierarchical architectures. More specifically, we first ablated the experiments for our HSC method without using the middle-level semantic feature $\mathbf{f}$ (denoted by Ours \textit{w./o.} FCN). We also ablated the setting without using the core semantics $\bm{\mathcal{S}}$ (denoted by Ours \textit{w./o.} SCN). {We then ablated the channel-wise setting by using an autogressive pixel-wise entropy model \cite{minnen2018joint} instead (denoted by Ours \textit{w./o.} CW) and further evaluated our HSC method without the mutual semantics between SCN and FCN (denoted by Ours \textit{w./o.} Hier).} We list the results in Table \ref{tab:ablation}. As can be seen from this table, without using FCN, the bit-rate can be reduced, whereas the restoration accuracy severely degraded. More importantly, using both SCN and hierarchical architectures can consistently improve the performance for all the three metrics. The above results verify the effectiveness of the core components within our HSC method. 

\begin{table}[ht]
\centering
\caption{Ablation study on the core components of our HSC method on the CelebA dataset, evaluated by the LPIPS, DISTS, and FID metrics. A downward arrow ($\downarrow$) indicates that a lower score is better for restoration accuracy and consistency. Bold font is used to highlight the best values.}
\label{tab:ablation}
\begin{tabular}{lllll}
\hline\hline
Ablation Settings& bpp    & LPIPS$\downarrow$ & DISTS$\downarrow$ & FID$\downarrow$   \\\hline
Ours \textit{w./o.} FCN     & 0.009 & 0.357 & 0.178 & 26.66 \\
Ours \textit{w./o.} SCN     & 0.009 & 0.359 & 0.173 & 37.77 \\
Ours \textit{w./o.} CW  & 0.012 & 0.265 & 0.132 & 13.14 \\
Ours \textit{w./o.} Hier    & 0.011 & \textbf{0.264} & 0.132 & 13.11 \\
Ours          & 0.011 & \textbf{0.264} & \textbf{0.131} & \textbf{12.93} \\
\hline\hline
\end{tabular}
\end{table}

\section{Conclusion}
In this paper, we have proposed a novel hierarchical semantic compression (HSC) framework that is able to restore semantic consistency of compressed images. The hierarchical framework was motived by our analysis on entropy relationships, taking advantages of our newly established semantic compression pipeline. Then, we have proposed the feature compression network (FCN) and semantic compression network (SCN), together with the semantic context model to hierarchically compress the middle-level semantic feature and core semantics, which were output from our general inversion encoder of StyleGAN. We have further introduced the corresponding editing pipeline to ensure the semantic consistency and editing realism. Experimental results have demonstrated the state-of-the-art performances for semantic compression, in terms of semantic consistency and restoration fidelity for human vision, as well as task accuracy for machine vision. We believe that the proposed HSC method could pave a new  semantic compression paradigm for both human and machine vision tasks. 

\small{
\bibliographystyle{IEEEtran}

\bibliography{reference}}

\begin{thebibliography}{10}
\providecommand{\url}[1]{#1}
\csname url@samestyle\endcsname
\providecommand{\newblock}{\relax}
\providecommand{\bibinfo}[2]{#2}
\providecommand{\BIBentrySTDinterwordspacing}{\spaceskip=0pt\relax}
\providecommand{\BIBentryALTinterwordstretchfactor}{4}
\providecommand{\BIBentryALTinterwordspacing}{\spaceskip=\fontdimen2\font plus
\BIBentryALTinterwordstretchfactor\fontdimen3\font minus \fontdimen4\font\relax}
\providecommand{\BIBforeignlanguage}[2]{{%
\expandafter\ifx\csname l@#1\endcsname\relax
\typeout{** WARNING: IEEEtran.bst: No hyphenation pattern has been}%
\typeout{** loaded for the language `#1'. Using the pattern for}%
\typeout{** the default language instead.}%
\else
\language=\csname l@#1\endcsname
\fi
#2}}
\providecommand{\BIBdecl}{\relax}
\BIBdecl

\bibitem{hudson2018jpeg}
G.~Hudson, A.~L{\'e}ger, B.~Niss, I.~Sebesty{\'e}n, and J.~Vaaben, ``Jpeg-1 standard 25 years: past, present, and future reasons for a success,'' \emph{Journal of Electronic Imaging}, vol.~27, no.~4, pp. 040\,901--040\,901, 2018.

\bibitem{wiegand2003overview}
T.~Wiegand, G.~J. Sullivan, G.~Bjontegaard, and A.~Luthra, ``Overview of the h. 264/avc video coding standard,'' \emph{IEEE Transactions on Circuits and Systems for Video Technology}, vol.~13, no.~7, pp. 560--576, 2003.

\bibitem{sullivan2012overview}
G.~Sullivan, J.~Ohm, W.~Han, and T.~Wiegand, ``Overview of the high efficiency video coding (hevc) standard,'' \emph{IEEE Transactions on Circuits and Systems for Video Technology}, vol.~22, no.~12, pp. 1649--1668, 2012.

\bibitem{bross2021overview}
B.~Bross, Y.~Wang, Y.~Ye, S.~Liu, J.~Chen, G.~Sullivan, and J.~Ohm, ``Overview of the versatile video coding ({VVC}) standard and its applications,'' \emph{IEEE Transactions on Circuits and Systems for Video Technology}, vol.~31, no.~10, pp. 3736--3764, 2021.

\bibitem{hussain2018image}
A.~J. Hussain, A.~Al-Fayadh, and N.~Radi, ``Image compression techniques: A survey in lossless and lossy algorithms,'' \emph{Neurocomputing}, vol. 300, pp. 44--69, 2018.

\bibitem{jamil2023learning}
S.~Jamil, M.~J. Piran, M.~Rahman, and O.-J. Kwon, ``Learning-driven lossy image compression: A comprehensive survey,'' \emph{Engineering Applications of Artificial Intelligence}, vol. 123, p. 106361, 2023.

\bibitem{balle2017end}
J.~Ball{\'e}, V.~Laparra, and E.~P. Simoncelli, ``End-to-end optimized image compression,'' in \emph{ICLR}, 2017.

\bibitem{balle2018variational}
J.~Ball{\'e}, D.~Minnen, S.~Singh, S.~Hwang, and N.~Johnston, ``Variational image compression with a scale hyperprior,'' \emph{arXiv preprint arXiv:1802.01436}, 2018.

\bibitem{cheng2020learned}
Z.~Cheng, H.~Sun, M.~Takeuchi, and J.~Katto, ``Learned image compression with discretized gaussian mixture likelihoods and attention modules,'' in \emph{IEEE/CVF CVPR}, 2020, pp. 7939--7948.

\bibitem{hu2021learning}
Y.~Hu, W.~Yang, Z.~Ma, and J.~Liu, ``Learning end-to-end lossy image compression: A benchmark,'' \emph{IEEE Transactions on Pattern Analysis and Machine Intelligence}, vol.~44, no.~8, pp. 4194--4211, 2021.

\bibitem{blau2019rethinking}
Y.~Blau and T.~Michaeli, ``Rethinking lossy compression: The rate-distortion-perception tradeoff,'' in \emph{ICML}.\hskip 1em plus 0.5em minus 0.4em\relax PMLR, 2019, pp. 675--685.

\bibitem{li2017closed}
S.~Li, M.~Xu, Y.~Ren, and Z.~Wang, ``Closed-form optimization on saliency-guided image compression for hevc-msp,'' \emph{IEEE Transactions on Multimedia}, vol.~20, no.~1, pp. 155--170, 2017.

\bibitem{jiang2022does}
L.~Jiang, Y.~Li, S.~Li, M.~Xu, S.~Lei, Y.~Guo, and B.~Huang, ``Does text attract attention on e-commerce images: A novel saliency prediction dataset and method,'' in \emph{IEEE/CVF CVPR}, 2022, pp. 2088--2097.

\bibitem{blau2018perception}
Y.~Blau and T.~Michaeli, ``The perception-distortion tradeoff,'' in \emph{IEEE/CVF CVPR}, 2018, pp. 6228--6237.

\bibitem{mentzer2020high}
F.~Mentzer, G.~D. Toderici, M.~Tschannen, and E.~Agustsson, ``High-fidelity generative image compression,'' \emph{Advances in NIPS}, vol.~33, pp. 11\,913--11\,924, 2020.

\bibitem{wang2021towards}
S.~Wang, S.~Wang, W.~Yang, X.~Zhang, S.~Wang, S.~Ma, and W.~Gao, ``Towards analysis-friendly face representation with scalable feature and texture compression,'' \emph{IEEE Transactions on Multimedia}, vol.~24, pp. 3169--3181, 2021.

\bibitem{arikan2024semantic}
L.~Arikan and T.~Weissman, ``Semantic image compression using textual transforms,'' in \emph{IEEE ISIT-W}.\hskip 1em plus 0.5em minus 0.4em\relax IEEE, 2024, pp. 1--6.

\bibitem{prakash2017semantic}
A.~Prakash, N.~Moran, S.~Garber, A.~DiLillo, and J.~Storer, ``Semantic perceptual image compression using deep convolution networks,'' in \emph{IEEE DCC}.\hskip 1em plus 0.5em minus 0.4em\relax IEEE, 2017, pp. 250--259.

\bibitem{yan2021sssic}
N.~Yan, C.~Gao, D.~Liu, H.~Li, L.~Li, and F.~Wu, ``Sssic: semantics-to-signal scalable image coding with learned structural representations,'' \emph{IEEE Transactions on Image Processing}, vol.~30, pp. 8939--8954, 2021.

\bibitem{shen2020interpreting}
Y.~Shen, J.~Gu, X.~Tang, and B.~Zhou, ``Interpreting the latent space of gans for semantic face editing,'' in \emph{IEEE/CVF CVPR}, 2020, pp. 9243--9252.

\bibitem{kwon2022diffusion}
M.~Kwon, J.~Jeong, and Y.~Uh, ``Diffusion models already have a semantic latent space,'' \emph{arXiv preprint arXiv:2210.10960}, 2022.

\bibitem{mao2023scalable}
Q.~Mao, C.~Wang, M.~Wang, S.~Wang, R.~Chen, L.~Jin, and S.~Ma, ``Scalable face image coding via stylegan prior: Towards compression for human-machine collaborative vision,'' \emph{IEEE Transactions on Image Processing}, 2023.

\bibitem{wang2021HFGI}
T.~Wang, Y.~Zhang, Y.~Fan, J.~Wang, and Q.~Chen, ``High-fidelity gan inversion for image attribute editing,'' in \emph{IEEE/CVF CVPR}, 2022.

\bibitem{he2022elic}
D.~He, Z.~Yang, W.~Peng, R.~Ma, H.~Qin, and Y.~Wang, ``Elic: Efficient learned image compression with unevenly grouped space-channel contextual adaptive coding,'' in \emph{IEEE/CVF CVPR}, 2022, pp. 5718--5727.

\bibitem{muckley2023improving}
M.~J. Muckley, A.~El-Nouby, K.~Ullrich, H.~J{\'e}gou, and J.~Verbeek, ``Improving statistical fidelity for neural image compression with implicit local likelihood models,'' in \emph{ICML}.\hskip 1em plus 0.5em minus 0.4em\relax PMLR, 2023, pp. 25\,426--25\,443.

\bibitem{lei2023text+sketch}
E.~Lei, Y.~B. Uslu, H.~Hassani, and S.~S. Bidokhti, ``Text+ sketch: Image compression at ultra low rates,'' in \emph{ICML 2023 Workshop on Neural Compression: From Information Theory to Applications}, 2023.

\bibitem{chang2022conceptual}
J.~Chang, Z.~Zhao, C.~Jia, S.~Wang, L.~Yang, Q.~Mao, J.~Zhang, and S.~Ma, ``Conceptual compression via deep structure and texture synthesis,'' \emph{IEEE Transactions on Image Processing}, vol.~31, pp. 2809--2823, 2022.

\bibitem{yao2022style}
X.~Yao, A.~Newson, Y.~Gousseau, and P.~Hellier, ``A style-based gan encoder for high fidelity reconstruction of images and videos,'' in \emph{European conference on computer vision}.\hskip 1em plus 0.5em minus 0.4em\relax Springer, 2022, pp. 581--597.

\bibitem{marcellin2000overview}
M.~W. Marcellin, M.~J. Gormish, A.~Bilgin, and M.~P. Boliek, ``An overview of jpeg-2000,'' in \emph{IEEE DCC}.\hskip 1em plus 0.5em minus 0.4em\relax IEEE, 2000, pp. 523--541.

\bibitem{toderici2015variable}
G.~Toderici, S.~M. O'Malley, S.~J. Hwang, D.~Vincent, D.~Minnen, S.~Baluja, M.~Covell, and R.~Sukthankar, ``Variable rate image compression with recurrent neural networks,'' \emph{arXiv preprint arXiv:1511.06085}, 2015.

\bibitem{minnen2018joint}
D.~Minnen, J.~Ball{\'e}, and G.~D. Toderici, ``Joint autoregressive and hierarchical priors for learned image compression,'' \emph{Advances in NIPS}, vol.~31, 2018.

\bibitem{xie2021enhanced}
Y.~Xie, K.~L. Cheng, and Q.~Chen, ``Enhanced invertible encoding for learned image compression,'' in \emph{Proceedings of the 29th ACM international conference on multimedia}, 2021, pp. 162--170.

\bibitem{wang2004image}
Z.~Wang, A.~C. Bovik, H.~R. Sheikh, and E.~P. Simoncelli, ``Image quality assessment: from error visibility to structural similarity,'' \emph{IEEE Transactions on Image Processing}, vol.~13, no.~4, pp. 600--612, 2004.

\bibitem{wang2003multiscale}
Z.~Wang, E.~P. Simoncelli, and A.~C. Bovik, ``Multiscale structural similarity for image quality assessment,'' in \emph{The Thrity-Seventh Asilomar Conference on Signals, Systems \& Computers, 2003}, vol.~2.\hskip 1em plus 0.5em minus 0.4em\relax Ieee, 2003, pp. 1398--1402.

\bibitem{kim2017deep}
J.~Kim and S.~Lee, ``Deep learning of human visual sensitivity in image quality assessment framework,'' in \emph{IEEE/CVF CVPR}, 2017, pp. 1676--1684.

\bibitem{rippel2017real}
O.~Rippel and L.~Bourdev, ``Real-time adaptive image compression,'' in \emph{ICML}.\hskip 1em plus 0.5em minus 0.4em\relax PMLR, 2017, pp. 2922--2930.

\bibitem{agustsson2019generative}
E.~Agustsson, M.~Tschannen, F.~Mentzer, R.~Timofte, and L.~Gool, ``Generative adversarial networks for extreme learned image compression,'' in \emph{IEEE/CVF CVPR}, 2019, pp. 221--231.

\bibitem{he2022po}
D.~He, Z.~Yang, H.~Yu, T.~Xu, J.~Luo, Y.~Chen, C.~Gao, X.~Shi, H.~Qin, and Y.~Wang, ``Po-elic: Perception-oriented efficient learned image coding,'' in \emph{IEEE/CVF CVPR}, 2022, pp. 1764--1769.

\bibitem{agustsson2023multi}
E.~Agustsson, D.~Minnen, G.~Toderici, and F.~Mentzer, ``Multi-realism image compression with a conditional generator,'' in \emph{IEEE/CVF CVPR}, 2023, pp. 22\,324--22\,333.

\bibitem{ho2020denoising}
J.~Ho, A.~Jain, and P.~Abbeel, ``Denoising diffusion probabilistic models,'' \emph{Advances in NIPS}, vol.~33, pp. 6840--6851, 2020.

\bibitem{song2020denoising}
J.~Song, C.~Meng, and S.~Ermon, ``Denoising diffusion implicit models,'' \emph{arXiv preprint arXiv:2010.02502}, 2020.

\bibitem{preechakul2022diffusion}
K.~Preechakul, N.~Chatthee, S.~Wizadwongsa, and S.~Suwajanakorn, ``Diffusion autoencoders: Toward a meaningful and decodable representation,'' in \emph{IEEE/CVF CVPR}, 2022, pp. 10\,619--10\,629.

\bibitem{yang2024lossy}
R.~Yang and S.~Mandt, ``Lossy image compression with conditional diffusion models,'' \emph{Advances in NIPS}, vol.~36, 2024.

\bibitem{zhang2024machine}
Y.~Zhang, C.~Jia, J.~Chang, and S.~Ma, ``Machine perception-driven facial image compression: A layered generative approach,'' \emph{IEEE Transactions on Circuits and Systems for Video Technology}, 2024.

\bibitem{huang2021deep}
D.~Huang, X.~Tao, F.~Gao, and J.~Lu, ``Deep learning-based image semantic coding for semantic communications,'' in \emph{IEEE GLOBECOM}.\hskip 1em plus 0.5em minus 0.4em\relax IEEE, 2021, pp. 1--6.

\bibitem{korber2024egic}
N.~K{\"o}rber, E.~Kromer, A.~Siebert, S.~Hauke, D.~Mueller-Gritschneder, and B.~Schuller, ``Egic: enhanced low-bit-rate generative image compression guided by semantic segmentation,'' in \emph{European Conference on Computer Vision}.\hskip 1em plus 0.5em minus 0.4em\relax Springer, 2024, pp. 202--220.

\bibitem{akbari2019dsslic}
M.~Akbari, J.~Liang, and J.~Han, ``Dsslic: {D}eep semantic segmentation-based layered image compression,'' in \emph{IEEE ICASSP}.\hskip 1em plus 0.5em minus 0.4em\relax IEEE, 2019, pp. 2042--2046.

\bibitem{chang2023semantic}
J.~Chang, J.~Zhang, J.~Li, S.~Wang, Q.~Mao, C.~Jia, S.~Ma, and W.~Gao, ``Semantic-aware visual decomposition for image coding,'' \emph{International Journal of Computer Vision}, vol. 131, no.~9, pp. 2333--2355, 2023.

\bibitem{lee2024neural}
H.~Lee, M.~Kim, J.-H. Kim, S.~Kim, D.~Oh, and J.~Lee, ``Neural image compression with text-guided encoding for both pixel-level and perceptual fidelity,'' \emph{arXiv preprint arXiv:2403.02944}, 2024.

\bibitem{li2021cross}
J.~Li, C.~Jia, X.~Zhang, S.~Ma, and W.~Gao, ``Cross modal compression: Towards human-comprehensible semantic compression,'' in \emph{Proceedings of the 29th ACM international conference on multimedia}, 2021, pp. 4230--4238.

\bibitem{careil2023towards}
M.~Careil, M.~J. Muckley, J.~Verbeek, and S.~Lathuili{\`e}re, ``Towards image compression with perfect realism at ultra-low bitrates,'' in \emph{The Twelfth International Conference on Learning Representations}, 2023.

\bibitem{li2024misc}
C.~Li, G.~Lu, D.~Feng, H.~Wu, Z.~Zhang, X.~Liu, G.~Zhai, W.~Lin, and W.~Zhang, ``‘misc: Ultra-low bitrate image semantic compression driven by large multimodal model,'' \emph{arXiv preprint arXiv:2402.16749}, 2024.

\bibitem{wu2023latent}
C.~H. Wu and F.~De~la Torre, ``A latent space of stochastic diffusion models for zero-shot image editing and guidance,'' in \emph{IEEE/CVF ICCV}, 2023, pp. 7378--7387.

\bibitem{guo2024smooth}
J.~Guo, X.~Xu, Y.~Pu, Z.~Ni, C.~Wang, M.~Vasu, S.~Song, G.~Huang, and H.~Shi, ``Smooth diffusion: Crafting smooth latent spaces in diffusion models,'' in \emph{IEEE/CVF CVPR}, 2024, pp. 7548--7558.

\bibitem{karras2020analyzing}
T.~Karras, S.~Laine, M.~Aittala, J.~Hellsten, J.~Lehtinen, and T.~Aila, ``Analyzing and improving the image quality of stylegan,'' in \emph{IEEE/CVF CVPR}, 2020, pp. 8110--8119.

\bibitem{xia2022gan}
W.~Xia, Y.~Zhang, Y.~Yang, J.-H. Xue, B.~Zhou, and M.-H. Yang, ``Gan inversion: A survey,'' \emph{IEEE Transactions on Pattern Analysis and Machine Intelligence}, vol.~45, no.~3, pp. 3121--3138, 2022.

\bibitem{minnen2020channel}
D.~Minnen and S.~Singh, ``Channel-wise autoregressive entropy models for learned image compression,'' in \emph{IEEE ICIP}.\hskip 1em plus 0.5em minus 0.4em\relax IEEE, 2020, pp. 3339--3343.

\bibitem{krizhevsky2012imagenet}
A.~Krizhevsky, I.~Sutskever, and G.~E. Hinton, ``Imagenet classification with deep convolutional neural networks,'' \emph{Advances in NIPS}, vol.~25, 2012.

\bibitem{wei2022e2style}
T.~Wei, D.~Chen, W.~Zhou, J.~Liao, W.~Zhang, L.~Yuan, G.~Hua, and N.~Yu, ``E2style: Improve the efficiency and effectiveness of stylegan inversion,'' \emph{IEEE Transactions on Image Processing}, 2022.

\bibitem{deng2019arcface}
J.~Deng, J.~Guo, N.~Xue, and S.~Zafeiriou, ``Arcface: Additive angular margin loss for deep face recognition,'' in \emph{IEEE/CVF CVPR}, 2019, pp. 4690--4699.

\bibitem{CelebAMask-HQ}
C.-H. Lee, Z.~Liu, L.~Wu, and P.~Luo, ``Maskgan: Towards diverse and interactive facial image manipulation,'' in \emph{IEEE/CVF CVPR}, 2020.

\bibitem{harkonen2020ganspace}
E.~H{\"a}rk{\"o}nen, A.~Hertzmann, J.~Lehtinen, and S.~Paris, ``Ganspace: Discovering interpretable gan controls,'' \emph{Advances in NIPS}, vol.~33, pp. 9841--9850, 2020.

\bibitem{karras2019style}
T.~Karras, S.~Laine, and T.~Aila, ``A style-based generator architecture for generative adversarial networks,'' in \emph{IEEE/CVF CVPR}, 2019, pp. 4401--4410.

\bibitem{krause20133d}
J.~Krause, M.~Stark, J.~Deng, and L.~Fei-Fei, ``3d object representations for fine-grained categorization,'' in \emph{Proceedings of the IEEE international conference on computer vision workshops}, 2013, pp. 554--561.

\bibitem{yu15lsun}
F.~Yu, Y.~Zhang, S.~Song, A.~Seff, and J.~Xiao, ``Lsun: Construction of a large-scale image dataset using deep learning with humans in the loop,'' \emph{arXiv preprint arXiv:1506.03365}, 2015.

\bibitem{Karras2017ProgressiveGO}
\BIBentryALTinterwordspacing
T.~Karras, T.~Aila, S.~Laine, and J.~Lehtinen, ``Progressive growing of gans for improved quality, stability, and variation,'' \emph{ArXiv}, vol. abs/1710.10196, 2017. [Online]. Available: \url{https://api.semanticscholar.org/CorpusID:3568073}
\BIBentrySTDinterwordspacing

\bibitem{zhang2018unreasonable}
R.~Zhang, P.~Isola, A.~A. Efros, E.~Shechtman, and O.~Wang, ``The unreasonable effectiveness of deep features as a perceptual metric,'' in \emph{IEEE/CVF CVPR}, 2018, pp. 586--595.

\bibitem{ding2020image}
K.~Ding, K.~Ma, S.~Wang, and E.~P. Simoncelli, ``Image quality assessment: Unifying structure and texture similarity,'' \emph{IEEE Transactions on Pattern Analysis and Machine Intelligence}, vol.~44, no.~5, pp. 2567--2581, 2020.

\bibitem{heusel2017gans}
M.~Heusel, H.~Ramsauer, T.~Unterthiner, B.~Nessler, and S.~Hochreiter, ``Gans trained by a two time-scale update rule converge to a local nash equilibrium,'' \emph{Advances in NIPS}, vol.~30, 2017.

\bibitem{garcia2017review}
A.~Garcia-Garcia, S.~Orts-Escolano, S.~Oprea, V.~Villena-Martinez, and J.~Garcia-Rodriguez, ``A review on deep learning techniques applied to semantic segmentation,'' \emph{arXiv preprint arXiv:1704.06857}, 2017.

\bibitem{liu2019variance}
L.~Liu, H.~Jiang, P.~He, W.~Chen, X.~Liu, J.~Gao, and J.~Han, ``On the variance of the adaptive learning rate and beyond. arxiv 2019,'' \emph{arXiv preprint arXiv:1908.03265}, 2019.

\bibitem{zhang2019lookahead}
M.~Zhang, J.~Lucas, J.~Ba, and G.~E. Hinton, ``Lookahead optimizer: k steps forward, 1 step back,'' \emph{Advances in NIPS}, vol.~32, 2019.

\bibitem{korber2024perco}
N.~K{\"o}rber, ``Perco (sd): Open perceptual compression,'' in \emph{Workshop on Machine Learning and Compression, NIPS 2024}.

\end{thebibliography}

\end{document}